\documentclass[journal,twoside,web]{IEEEtran}

\usepackage{textgreek}
\usepackage{tabto}
\usepackage{tabu}
\usepackage{array, booktabs, makecell}
\usepackage{float}
\usepackage{xcolor}
\usepackage{colortbl}
\usepackage{commath}
\usepackage{multirow}
\usepackage{amsmath}
\usepackage{amsfonts}
\usepackage{enumitem}
\usepackage{nicefrac}
\usepackage{wrapfig}

\usepackage{algorithm}
\usepackage{algpseudocode}

\usepackage{xcolor}
\usepackage{academicons}
\usepackage{tikz,pgfplots,adjustbox}
\usepackage{filecontents}
\usepackage{tabularx}
\usepackage{pgfplotstable}
\definecolor{orcidlogocol}{HTML}{A6CE39}

\usepackage{overpic}

\usepackage{hyperref}

\newcolumntype{Y}{>{\centering\arraybackslash}X}
\usepackage{hyperref}

\newcolumntype{Y}{>{\centering\arraybackslash}X}

\RequirePackage{pgf}
\RequirePackage{pgffor}
\RequirePackage{pgfpages}
\RequirePackage{tikz} 
\usetikzlibrary{
	calc,
	arrows,
	arrows.meta,
	positioning,
	matrix,
	shapes,
	shapes.geometric,
	shapes.misc,
	shapes.callouts,
	shapes.arrows,
	scopes,
	automata,
	shadows,
	fit,
	petri,
	backgrounds,
	decorations.pathreplacing,
	decorations.pathmorphing,
	decorations.markings,
	fadings,
	spy, 
	patterns
}

\usepackage{pifont}

\RequirePackage[acronym,shortcuts]{glossaries}
\glsdisablehyper

\newacronym[plural=ADs,firstplural={aortic dissections (ADs)}]{AD}{AD}{aortic dissection}
\newacronym{CTA}{CTA}{computed tomography angiography}
\newacronym{FL}{FL}{false lumen}
\newacronym{TL}{TL}{true lumen}
\newacronym{TAAD}{TAAD}{type~A aortic dissection}
\newacronym{TBAD}{TBAD}{type~B aortic dissection}
\newacronym{GAN}{GAN}{generative adversarial network}
\newacronym{LOA}{LOA}{limits of agreement}
\newacronym[plural=CNN, firstplural={convolutional neural networks (CNN)}]{CNN}{CNN}{convolutional neural network}
\newacronym{MCDS}{MCDS}{Monte Carlo dropout sampling}
\newacronym{MCDbS}{MCDbS}{Monte Carlo DropBlock sampling}
\newacronym{INS}{INS}{iterative neighbor sampling}
\newacronym{INDS}{INDS}{iterative neighbor dropout sampling}

\usepackage{cite}
\usepackage{amsmath,amssymb,amsfonts}
\usepackage{graphicx}
\usepackage{textcomp}

\def\BibTeX{{\rm B\kern-.05em{\sc i\kern-.025em b}\kern-.08em
    T\kern-.1667em\lower.7ex\hbox{E}\kern-.125emX}}

\ifCLASSINFOpdf
\else
\fi

\begin{document}


%
\title{Deep Medial Voxels: Learned Medial Axis Approximations for Anatomical Shape Modeling}



\author{Antonio Pepe, Richard Schussnig, Jianning Li, 
Christina Gsaxner, Dieter Schmalstieg, \IEEEmembership{Fellow, IEEE}, \\ and Jan Egger
\thanks{Manuscript received XXX; revised XXX;
accepted XXX. Date of publication XXX;
date of current version XXX. 
This work received funding from the TU Graz LEAD Project \emph{Mechanics, Modeling and Simulation of Aortic Dissection}. (Corresponding author: Antonio Pepe.)}
\thanks{A. Pepe, C. Gsaxner, J. Li, D. Schmalstieg and J. Egger are with the Graz University of Technology, Institute of Computer Graphics and Vision, Inffeldgasse 16a, 8010 Graz, Austria. (e-mail: \{antonio.pepe, gsaxner, jianning.li, schmalstieg, egger\}@tugraz.at).}
\thanks{R. Schussnig is with the Ruhr University Bochum, Faculty of Mathematics, 44780, Bochum, Germany. (e-mail: richard.schussnig@rub.de).}
\thanks{J. Li and J. Egger are also with the Essen University Hospital, Institute for Artificial Intelligence in Medicine, Girardetstraße 2, 45131 Essen, Germany.}
\thanks{D. Schmalstieg is also with the University of Stuttgart, Stuttgart, Germany.}
}

\maketitle

\begin{abstract}
Shape reconstruction from imaging volumes is a recurring need in medical image analysis. Common workflows start with a segmentation step, followed by careful post-processing and, finally, ad hoc meshing algorithms. As this sequence can be time-consuming, neural networks are trained to reconstruct shapes through template deformation. These networks deliver state-of-the-art results without manual intervention, but, so far, they have primarily been evaluated on anatomical shapes with little topological variety between individuals. In contrast, other works favor learning implicit shape models, which have multiple benefits for meshing and visualization. Our work follows this direction by introducing deep medial voxels, a semi-implicit representation that faithfully approximates the topological skeleton from imaging volumes and eventually leads to shape reconstruction via convolution surfaces. Our reconstruction technique shows potential for both visualization and computer simulations. 
\end{abstract}

\begin{IEEEkeywords}
convolutional neural networks, convolution surface, distance function, centerline, medial axis transform, implicit model
\end{IEEEkeywords}

\section{Introduction}
\label{sec:introduction}

\glsresetall
\IEEEPARstart{T}{he} reconstruction of anatomical regions is a prerequisite for numerous tasks in medical image analysis, ranging from visualization~\cite{mistelbauer2021implicit,oeltze2005visualization} to implant design~\cite{li2021autoimplant} and physics-based simulations of pathological processes~\cite{baumler2020fluid,shad2021patient}. Traditionally, this is performed after the initial binary segmentation of an imaging volume~\cite{li2021autoimplant, mistelbauer2021implicit}. With an emphasis on blood vessels, different studies and challenges analyzed the need for a smooth, organic-looking and artifact-free representation in clinical  visualizations~\cite{oeltze2005visualization, mistelbauer2021implicit} and patient-specific computer simulations~\cite{antiga2008image, kerrien2017blood, mistelbauer2021implicit,bovsnjak2023higher}. The recent SEG.A. challenge~\footnote{SEG.A. 2023~\url{https://multicenteraorta.grand-challenge.org/}} showed that the quantitatively best-performing algorithms were not necessarily the solution of choice for clinicians and biomedical simulations. A further shared output of recent studies is that \emph{skeleton-based} implicit representations play a key role in both domains. Implicit representations describe shapes analytically in the form of level sets, while the skeleton allows one to quantify vascular growth~\cite{burris2022vascular} but also to parametrically\footnote{A parametric representation is here considered as a lower-dimensional description of a surface, e.g., skeleton-based parametrization.} model and mesh the geometry~\cite{mistelbauer2021implicit,bovsnjak2023higher} for use cases like in-silico trials~\cite{sarrami2021silico}. In these studies, the skeletons were simply represented as centerlines and first required binary segmentation.

Recent work showed that discretized distance functions, a type of implicit representation, can be learned by neural networks at least as efficiently as binary representations for different organs~\cite{wang2020deep,li2020shape}. Less attention has been paid to learning implicit representations based on skeletal structures such as the medial axis transform (MAT)~\cite{hu2019mat,li2020shape,lin2021point2skeleton}. In this paper, we introduce \emph{deep medial voxels} (DMV), the first effort to create parametric meshes directly from imaging volumes, with several advantages:

\begin{enumerate}
\item No explicit binary segmentation step is required; instead, a segmentation of the underlying geometry is intrinsically provided as a level-set representation.
\item We generate a skeleton mesh and apply surface convolution~\cite{suarez2019anisotropic} to obtain a surface mesh. Due to the parameterization provided by the skeleton, the surface is smooth, watertight and unaffected by voxel discretization. 
\item A differentiable formulation of a shape skeleton, which allows one to analyze topology information during training and back-propagate a topological loss.\\

\hspace{-6mm} Moreover, DMV allows to
    \item Co-segment the reconstructed shape by clustering the predicted skeleton
    \item Identify in- and outflow interfaces for computer simulation of vasculature by analyzing the skeleton connectivity
    \item Produce meshes that can be readily applied in numerical simulations
\end{enumerate}

\begin{figure*}[t]
    \centering
    \includegraphics[width=\linewidth]{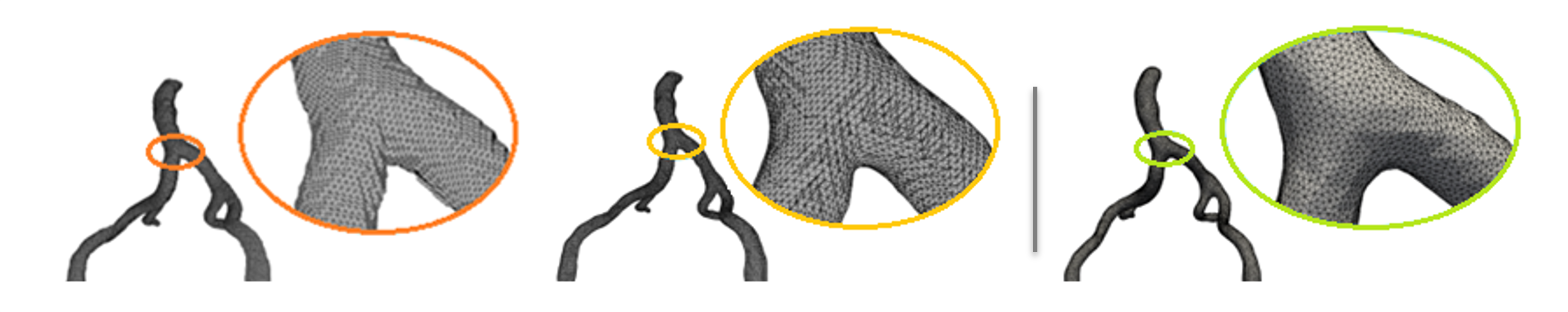}
    \caption{A surface mesh of an iliac bifurcation on the aorta is obtained (left) directly from a binary segmentation using marching cubes, (middle) from a discrete distance field generated from the binary segmentation. In both cases, voxel artifacts can be noticed. (right) The convolution surface of the medial voxels obtained with differentiable DMV approach. }
    \label{fig:mc_vs_conv}
\end{figure*}

\section{Background}
\label{sec:background}

Conventionally, polygonal surface meshes of segmented volumes are generated with algorithms such as marching cubes~\cite{li2021autoimplant,newman2006survey} and often suffer from discretization artifacts (\autoref{fig:mc_vs_conv})~\cite{hong2019high,kerrien2017blood} that limit their use for visualization~\cite{li2021autoimplant,oeltze2005visualization} and simulation~\cite{kerrien2017blood}. Various meshing and editing methods have been designed to overcome such limitations. Some of them were integrated into software such as SimVascular and VMTK~\cite{baumler2020fluid,piccinelli2009framework}. A recent trend is to model anatomical shapes by learning differentiable prior deformations~\cite{wickramasinghe2020voxel2mesh,pak2021distortion}, bypassing binary segmentation steps without losing accuracy. Yet, such learned mesh deformations operate at a fixed resolution and require a shape prior. As an alternative, we rely on implicit representations, which have been shown to provide optimal quality in both visualization and simulation applications~\cite{mistelbauer2021implicit,antiga2008image}. Here, we provide an overview of implicit shape modeling, introduce the medial axis transform as a shape description tool, and discuss how our work is related to the current state of the art in learned shape modeling. 

\begin{figure}
    \centering
    \includegraphics[width=\columnwidth]{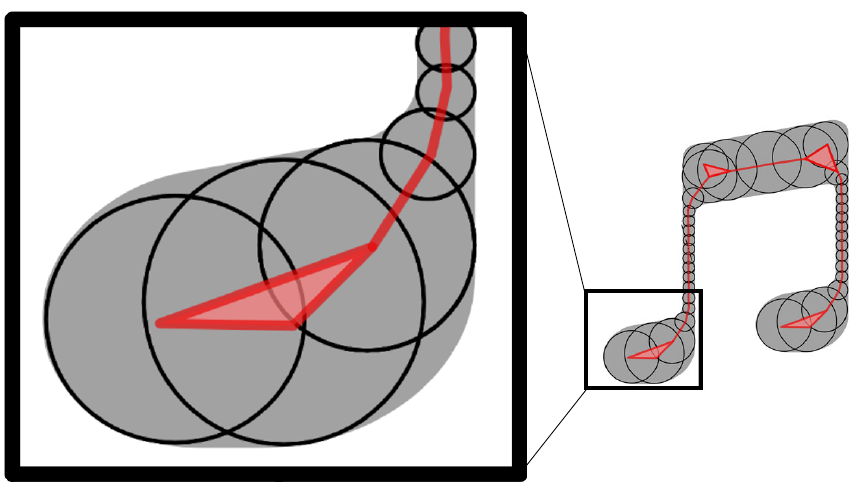}
    \caption{A simplicial complex (red) over a 2D shape (gray). The black circles are a representative subset of the medial axis transform. If two circles overlap, their centers are connected by a line. If three circles overlaps, their centers are connected by a triangle. The same concept holds for 3D shapes where circles are replaced by spheres.}
    \label{fig:simp_complex}
\end{figure}

\subsection{Implicit shape modeling}
\emph{Implicit representations} describe shapes as a level set of a function $f(P):\mathbb{R}^3\rightarrow\mathbb{R}$~\cite{hong2019high} and a parameter $C\in\mathbb{R}$:
\begin{equation}
    f(P) - C = 0.
    \label{eq:implicit}
\end{equation}
Examples are distance transforms~\cite{wang2020deep}, metaballs~\cite{kanamori2008gpu}, and convolution surfaces~\cite{oeltze2005visualization}. Their analytical formulation allows the extraction of high-resolution meshes without discretization artifacts~\cite{fries2017higher,oeltze2005visualization}. They can easily model branching structures and ensure smooth surfaces by relying on convolution operations. Given a skeletal structure $\Gamma~=~\bigcup_i~\Gamma_i$, described as the union of a set of primitives (e.g., polylines) with thickness $R_i$, a convolution surface is defined as
\begin{equation}
    \label{eq:conv_surf_part}
    f(P) = \sum_i \int_{\Gamma_i} e^{-k\frac{\left|\left|P - s\right|\right|_2^2}{R_i^2}} ds,
\end{equation}
where the convolution kernel (e.g.,  $e^{-kx}$) is $\mathcal{C}^0$ and strictly decreasing~\cite{oeltze2005visualization, suarez2019anisotropic}. A more detailed description of convolution surfaces was recently provided by Suárez et al.~\cite{suarez2019anisotropic}. 

Here, we use skeletons represented as a \emph{simplicial complex} (\autoref{fig:simp_complex}), a topological representation that is gaining popularity in different fields~\cite{torres2020simplicial}. A simplicial complex is a non-manifold mesh formed by a set of vertices, lines, and triangles. Each subset of $n+1$ interconnected vertices forms an $n$-dimensional simplex and is the basic unit of the complex~\cite{edelsbrunner2022computational}. 

\begin{figure*}
    \centering
    \includegraphics[width=0.95\linewidth]{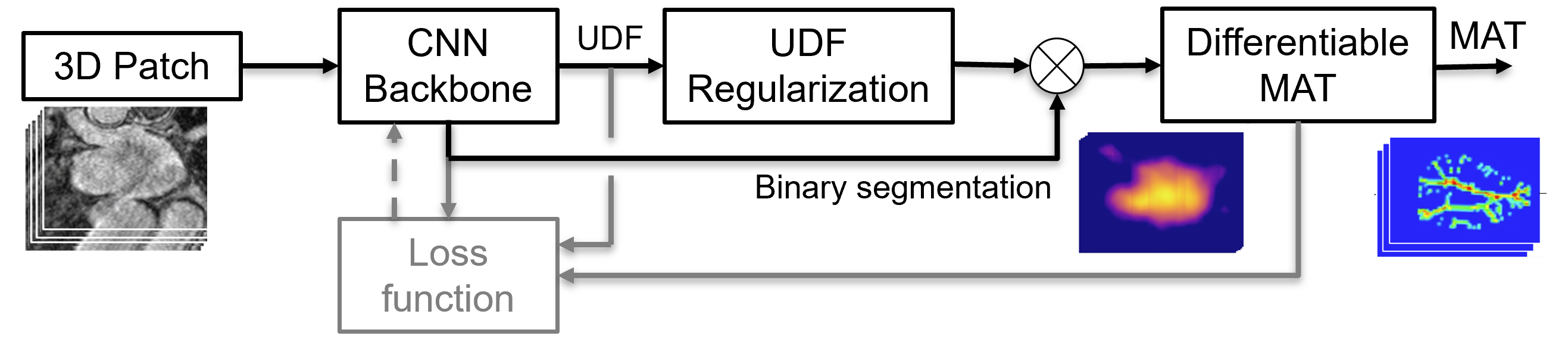}
    \caption{Illustration of our pipeline. A CNN backbone is used to predict the binary segmentation and the UDF of the target organ. The UDF is regularized to ensure local continuity and a differentiable MAT is extracted, which provides topology information in the loss function. The gray steps are only executed during training.}
    \label{fig:arch}
\end{figure*}

\subsection{Medial axis transform} 

Centerlines are common for the topological analysis of tubular shapes~\cite{mistelbauer2021implicit,oeltze2005visualization,wang2020deep}. The \textit{medial axis transform} (MAT) can be thought of as a generalization of the centerlines to generic shapes. The MAT of a closed surface $S$ is defined as the set of all maximally inscribed spheres such that each sphere is tangent to $S$ in at least two points~\cite{giesen2009scale}. However, even small amounts of noise can cause large variations in the MAT~\cite{li2015q}. Various studies have investigated more stable and simpler approximations of the MAT~\cite{dou2021coverage,giesen2009scale,li2015q,yan2018voxel}. An early approximation, $\lambda$-MAT, considers only spheres with radius $r\geq\lambda$. Its drawback is that topological changes are introduced if regions with a maximum medial radius $r_M \le \lambda$ are present~\cite{giesen2009scale}. 

The \emph{scale axis transform} considers exactly those spheres which are centered on the topological skeleton ~\cite{giesen2009scale}. Another robust representation is \emph{LS-MAT}~\cite{rebain2019lsmat}, which estimates a set of MAT spheres through a least-squares optimization. \emph{Q-MAT} approximates the MAT through a \textit{simplicial complex}~\cite{li2015q} with a user-defined number of vertices. Each simplex, line or triangle models a truncated cone or a medial slab that approximates the volume of the removed medial spheres that were centered along the simplex. 

Skeletal structures based on simplicial complexes not only allow to describe shapes with a limited amount of parameters, but they have also proven to be useful in shape co-segmentation~\cite{lin2020seg} and landmark detection~\cite{lin2021point2skeleton}. \emph{Voxel cores}~\cite{yan2018voxel} define a MAT over a binary voxel grid by its distance transform. The distance value $d_i$ of a voxel $v_i$ represents the radius of the medial sphere centered in $v_i$. A threshold is used to prune all spheres with radius $r \leq \lambda$ that do not lead to a topological change. Here, we extend these notions and develop a differentiable approximation of the MAT for a topology-aware training of neural networks. 

\subsection{Learning shape representations}

Implicit neural representations were introduced to encode a given geometric shape within the weights of a neural network~\cite{park2019deepsdf, rebain2021deep} or to infer a parametric description of a given mesh~\cite{genova2020local}. Point2Skeleton~\cite{lin2021point2skeleton} and P2MAT-Net~\cite{yang2020p2mat} infer a MAT approximation from a sparse surface point cloud. Their output is represented as a simplicial complex, which allows reconstructing surfaces as the union of medial primitives~\cite{li2015q}. For shape \emph{completion}, such methods show advantages over the common Poisson reconstruction~\cite{erler2020points2surf}. Learning strategies have also been suggested for the reconstruction of shapes from volume grids~\cite{peng2020convolutional}. Volume grids are used extensively in medical imaging. For shapes such as the hippocampus~\cite{wickramasinghe2020voxel2mesh}, learned template deformations generate explicit meshes from imaging volumes~\cite{wickramasinghe2020voxel2mesh, pak2021distortion, zhao2022segmentation}. 

Learning to infer implicit representations from such volumes has received less attention, although some literature suggests a high potential~\cite{ma2020distance}. An example is \emph{Deep Distance Transform} (DDT)~\cite{wang2020deep}, which infers the distance transform of \emph{tubular} shapes, like the aorta. Li et al.~\cite{li2020shape} learn a normalized distance function from MRI data to reconstruct the left cardiac atrium. They show that adversarial learning of distance functions can improve the generalization of the network. More recently, Lin et al.~\cite{lin2023structure} suggested a skeleton-aware distance transform for 2D instance segmentation by learning a weighted energy functional derived from the distance transform. Their work showed the importance of considering topological information when learning implicit representations. Zhang et al.~\cite{zhang2023anatomy} showed the importance of preserving topology when learning to segment tubular structures such as coronary arteries. However, as remarked by Lin et al.~\cite{lin2023structure}, none of these methods infers a parametric 3D representation, and all are limited to a specific organ. We fill this gap by learning differentiable MAT approximations from volumes and reconstruct the original shape by means of convolution surfaces. 


\begin{figure*}
    \centering
    \includegraphics[width=0.95\linewidth]{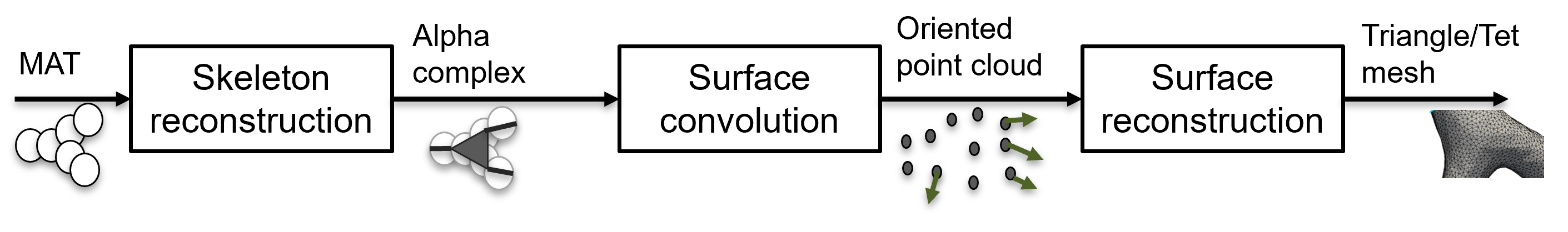}
    \caption{Following the prediction of the MAT as shown in \autoref{fig:arch}, we convert the MAT to a weighted alpha complex. The convolution surface of the complex provides an oriented point cloud of the surface shape. This point cloud allows reconstructing both triangular and tetrahedral mesh representations.}
    \label{fig:arch2}
\end{figure*}

\section{Computing MAT from imaging volume}
In the first stage, our method accepts a raw voxel grid as input and automatically computes an approximate MAT of the target shape represented in the input volume. The algorithm approximates the MAT with configurable fidelity, supporting the extraction of meshes with different levels of detail.

\subsection{CNN backbone and UDF regularization} 
\label{sec:backbone}

In a supervised setting, a U-Net-like CNN backbone is trained to jointly infer binary segmentation and UDF from a raw imaging volume.  UDF levels are binned into $K$ classes, with class $0$ representing the background.

The CNN, fed with a 3D patch, is trained to predict the discretized UDF. In our implementation, given a volume $V$ delimited by the surface $S$ and a point $P = (x,y,z) \in \mathbb{R}^3$, we define the UDF as
\begin{equation}
  u(P) = \left\{\begin{array}{@{}l@{}}
   \,\, \lfloor \underset{k \in S}{\text{inf}} \| P - k \|_2+0.5\rfloor, \quad \mbox{if} \,\, P \in V\\
    \quad\quad    0, \quad\quad\quad\,\,\, \mbox{otherwise}.
  \end{array}\right.\,
\end{equation}
The UDF is then multiplied by the predicted binary segmentation to remove spurious noise. Now, for the definition of UDF, each voxel $v$ with distance $u(v)~>~0$ must have at least one neighbor $w\in\mathcal{N}(v)$ with distance $u(w) = u(v)-1$.
To guarantee spatial coherence in the prediction, we recompute the output as 
\begin{equation}
    u_q(v) = \min_{w \in \mathcal{N}(v)} (u(w)) + 1.
\end{equation}

\subsection{Differentiable relaxation of MAT}
\label{sec:diffMAT}

Unlike previous work~\cite{li2020shape,ma2020distance, wang2020deep, wickramasinghe2020voxel2mesh}, we promote the preservation of topology during training by introducing a differentiable approximation of MAT. The definition of MAT implies that the UDF will present a local maximum, or \emph{ridge}, at the center of each sphere, where the local UDF value is the radius of the sphere~\cite{xia2011fast}. However, depending on the resolution of the grid, local maxima may not be present for smaller structures or may be neglected by the CNN. Therefore, we introduce a relaxation of this definition and retrieve all MAT \emph{candidate points} that are locally dominant 
\begin{equation}
    \label{eq:dMAT}
    C_{rel} = \{ (P,r) \,\, : \,\, u(P) > \mu(\mathcal{N}_u(P)) \land r = u(P) \},
\end{equation}
where $\mu$ is the average UDF value in a neighborhood $\mathcal{N}_u$ of $P$. This definition includes all local maxima $C \subseteq C_{rel}$. Any non-maximally inscribed sphere will not negatively impact the shape reconstruction. To allow differentiation during neural network training, we approximate the inequality in \autoref{eq:dMAT} as
\begin{equation}
    C_{d} = \{ (P, r) \,\, : \,\, \sigma(u(P) - \mu(\mathcal{N}_u(P))) > 0 \land r = u(P) \},
\end{equation}
where $\sigma(x) = (1 + e^{-kx})^{-1}$ is the scaled Sigmoid function.
For $k=50$, $\sigma(x)$ provides a good differentiable approximation of the step function~\cite{liao2020real}.

\subsection{Topology and ridge-preserving loss} 
\label{sec:loss}

Previous work focused only on distance values~\cite{li2020shape,wang2020deep} or, for tubular structures, on a soft centerline to preserve topology~\cite{shit2021cldice}. For more general shapes, we must preserve the entire MAT. 
\begin{equation}
\begin{split}
    \mathcal{L}(\hat{y}_s,\hat{y}_u,y_s, y_u) = l_s(\hat{y_s},y_s)  + l_u(\hat{y_u},y_u) + 
    \\
    + l_L(\hat{y}_u, y_u) + l_m(\hat{y}_u,y_u) 
\end{split}
\end{equation}
where $\hat{y}_s$ is the predicted segmentation, $\hat{y}_u$ is the UDF, and $y_s$ and $y_u$ are the respective ground truths. The function $l_s(\cdot,\cdot)$ is the Dice score similarity loss.
For the UDF, the function $l_u(\cdot, \cdot)$ is a stochastic distance loss
\begin{equation}
    l_u(\hat{y},y) = - \sum_{i=1}^{N} \sum_{k=1}^{K} \Big[ \mathbf{1}(\hat{y}_{s,i} = k) \cdot \mathcal{P}(\hat{y}_{s,i}=k \,\small|\, {y}_{s,i}) \Big],
\end{equation}
where $\bold{1}(\cdot, \cdot)$ is an indicator function, and $\mathcal{P}(\hat{y}_{s,i}=k \,\small|\, {y}_{s,i})$ is the conditional probability of $\hat{y}_{s,i}$, given $y_{s,i}$ and a normal distribution with variance $\sigma = 1$. To preserve UDF continuity, we minimize the $L_2$ distance 
\begin{equation}
    l_L(\hat{y}_u, y_u) = L_2(\nabla^2\hat{y}_u, \nabla^2 y_u)
\end{equation}
of the Laplacian 
\begin{equation}
    \nabla^2 = \frac{\partial^2 \cdot}{\partial x^2} + \frac{\partial^2 \cdot}{\partial y^2} + \frac{\partial^2 \cdot}{\partial z^2}.
\end{equation}
Finally, we evaluate the MAT candidate points to preserve the topology as
\begin{equation}
    l_m(\hat{y}_u, y_u) = l_u(C_d(\hat{y}_u), C_d(y_u)),
\end{equation}
which increases the UDF loss for the MAT points $C_d(\cdot)$, the topological skeleton. 

\section{Computation of mesh from MAT}

In the second part of our method, we use the predicted MAT to generate a skeleton and, from this skeleton, reconstruct the entire shape.

\subsection{Skeleton generation}
The MAT is a cloud of 4D points $(x,y,z,r)$, i.e., the ridges of the discrete UDF (\autoref{fig:arch2}). To generate a skeleton, different methodologies on how to link medial spheres have been suggested~\cite{lin2021point2skeleton}. The Nerve theorem guarantees that a cloud of spheres can be dually represented as a \v{C}hech complex, a simplicial complex where two overlapping spheres are connected by an edge, and, three, by a triangle~\cite{edelsbrunner2022computational}. As \v{C}hech complexes are computationally expensive, research focused on how to approximate them~\cite{dou2021coverage, lin2021point2skeleton, yang2020p2mat}. We approximate the \v{C}hech complex by building a \emph{Laguerre-Voronoi diagram} from the MAT points~\cite{dou2021coverage}. If the cells of three points are adjacent in pairs in the diagram, they form a triangle; two form a segment. This form leads to a simplicial complex known as a weighted alpha complex~\cite{edelsbrunner2022computational}. As the complete MAT can present a large number of elements, we rely on Q-MAT~\cite{li2015q} to simplify the simplicial complex if necessary.

\begin{figure*}
    \centering
    \includegraphics[width=\linewidth]{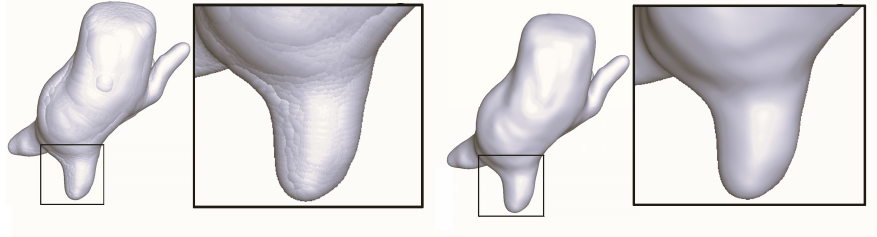}
    \caption{Shape reconstruction from the MAT is usually computed as the union of medial primitives~\cite{li2015q,lin2021point2skeleton}. (left) The union operation does not guarantee surface continuity and smoothness, leading to a series of ``bumps''. (right) Instead, we combine the MAT with a convolution surface. The convolution guarantees a smooth and watertight reconstruction.}
    \label{fig:unionVSdmv}
\end{figure*}
\subsection{Shape reconstruction as convolution surface} 
\label{sec:convolution}

The MAT and the skeleton provide a sparse and lightweight representation of the underlying shape. For shape reconstruction, we extend the notion of convolution surfaces to simplicial complexes. Convolution guarantees shape continuity of $\mathcal{C}^0$ and higher orders; this property improves visual quality~\cite{oeltze2005visualization} and is a frequent prerequisite for finite element analysis~\cite{fries2017higher,bovsnjak2023higher}. Convolution over lines and triangles requires different formulations of the partition integrals~\cite{mccormack1998creating} given in \autoref{eq:conv_surf_part}.

\subsubsection*{Segment convolution} 

A line segment $\mathbf{AB}$ with vertex radii $r_A,r_B\in\mathbb{R}^+$ can be described in the form $s(x) = A + lx\mathbf{u}$, where $l$ is its length and $\mathbf{u}$ its versor. The integration over the segments for a point $P\in\mathbb{R}^3$ is therefore
\begin{equation}
    f_{line}(P) = \int_0^1 K\Big( \lVert P - s(x)\rVert_2, r_{s}\Big) dx,
\end{equation}
where $K$ is the convolution kernel, and $r_s$ is the interpolation of the radii at $s(x)$.

\subsubsection*{Triangle convolution} 

A general triangle $T$ is split into two right-angled triangles $T_1$ and $T_2$. These triangles are computed independently as
\begin{equation}
    f_{triang}(P) = \int_0^{1} \int_{1}^{1-y} K\Big(\lVert P - P_{xy}\rVert_2,r_{xy}\Big) \,dx\,dy,
\end{equation}
where $P_{xy}\in\mathbb{R}^3$ is a point in the triangle with normalized projections $x$ and $y$ on the two catheti, and $r_{xy}\in\mathbb{R}$ is a bilinear interpolation of the radii at $P_{xy}$. Both cases rely on the kernel function $K(d,r) = \nicefrac{r^2}{(d^2+r^2})$. The desired level set has distance $d = r$ and, therefore, following \autoref{eq:implicit}, $C=K(r,r)=0.5$.

\subsubsection*{Mesh generation} 
The convolution surface definition allows us to derive explicit surface and volumetric mesh representations. We do this in two steps: 

1) We compute the convolution field on a volume grid. We extract the \emph{oriented point cloud} from the voxels of level set C, where the point normal is opposite to the convolution surface gradient. To account for grid discretization, we optimize the point location along its normal \label{sec:reconstruction}  according to the scheme of Algorithm~\autoref{algo:optim}, in which the hyperbolic tangent both modulates the step size and inverts the orientation if the surface level has been exceeded. 

\algnewcommand\algorithmicforeach{\textbf{for each:}}
\algnewcommand\ForEach{\item[ \algorithmicforeach]}

\algdef{SE}[DOWHILE]{Do}{doWhile}{\algorithmicdo}[1]{\,\,\,\,\,\,\algorithmicwhile\ #1}%
    
\begin{algorithm}
    \caption{Iterative point cloud refinement. \\ $\omega$: maximum tolerated error, $C$: surface level, $k$: maximum step size.}
    \begin{algorithmic} \label{algo:optim}
     \ForEach{cloud point P} 
     \State \textbf{N} = -$\frac{\partial \hat{S}(P)}{\partial P}$
     \Do :
     \State  $P = P + k \cdot \mbox{tanh}(\hat{S}(P) - C) \cdot \frac{\mathbf{N}}{\|\mathbf{N}\|}$       
     \doWhile {$\| \hat{S}(P) - C\|$ $\geq$ $\omega$}
    \end{algorithmic}
    \addtocounter{algorithm}{-1}
\end{algorithm}

2) We rely on local Poisson reconstruction for watertight surface triangulation of the oriented point cloud~\cite{kazhdan2006poisson}, optionally followed by volume tetrahedralization~\cite{baumler2020fluid,Geuzaine2009Gmsh}.   

\begin{table*}
\caption{Ablation study. The baseline consists of a V-Net~\cite{milletari2016v} backbone trained using the Dice score for the binary output and the mean squared error for the UDF output. All tests are evaluated on the left atrium dataset using a five-fold cross validation.}\label{tab2}
\centering
\begin{tabularx}{\linewidth}{|c|Y|Y|}
\hline
{\textbf{Method}} & {Dice Score \mbox{$\uparrow$[$\%$]}} &  {Hausdorff \,\,\,Distance \mbox{$\downarrow$[mm]}}  \\ 
\hline
Baseline (BL) &  85.53 $\pm$ 9.86 & 12.24 $\pm$ 6.01 \\ 
BL +  Laplacian Loss (LAPL)&  88.23 $\pm$ 8.67 &   8.01 $\pm$ 5.47 \\ 
BL + MAT Loss (MATL)& 89.80 $\pm$ 7.44  & 7.38 $\pm$ 5.06 \\ 
BL + MATL + LAPL  &  91.25 $\pm$ 4.29  & 5.47 $\pm$ 4.03 \\ 
\textbf{BL + MATL + LAPL + UDF Reg. (DMV)}  & \textbf{96.02} $\pm$ \textbf{3.20} &  \textbf{3.64} $\pm$ \textbf{2.93}  \\ 
\hline
\end{tabularx}
\end{table*}

\section{Experiments}

Following previous work~\cite{wang2020deep}, we carried out different experiments on four anatomical shapes and compared the results with four methods~\cite{li2020shape,milletari2016v,wang2020deep,wickramasinghe2020voxel2mesh} with a five-fold cross-validation. All experiments were developed using PyTorch v.1.8.0 on a desktop PC (CPU: Intel i7 3.20~GHz, 64 GB RAM, GPU: NVIDIA Titan RTX).

\begin{table*}
\caption{Comparison against state-of-the-art methods for different organs and modalities. AO: Aorta. AD: Aortic Dissection. LA: Cardiac Left Atrium. HI: Hippocampus.}\label{tab3}
\centering
\begin{tabularx}{.97\linewidth}{|c|X|X|X|X|X|X|} 
\hline
\multirow{2}{*}{\textbf{Method}} &  \multicolumn{3}{c|}{Dice Score \mbox{$\uparrow$[$\%$]}} &  \multicolumn{3}{c|}{Hausdorff Distance \mbox{$\downarrow$[mm]}} \\ 
   & AO, CT & AD, CT & LA, MRI & AO, CT & AD, CT & LA, MRI \\ 
\hline
V-Net~\cite{milletari2016v}  &  78.53$\pm$ 8.01 & 85.33 $\pm$ 6.54 & 89.01 $\pm$ 12.23 &  19.16$\pm$11.51 & 32.60 $\pm$ 20.15 & 11.34 ~$\pm$ 7.10  \\ 
SASSNet~\cite{li2020shape} & 82.84$\pm$6.12 & 89.08 $\pm$ 5.49 & 89.50 $\pm$ 9.00 &  18.21$\pm$ 10.60 & 21.60 $\pm$ 10.82 & 8.25 $\pm$ 5.22 \\ 
DDT ~\cite{wang2020deep}   & 90.83 $\pm$ 4.14 &  88.93 $\pm$ 4.19  & 88.53 $\pm$ 9.11 &  9.02 $\pm$ 3.76 & 8.12 $\pm$ 4.02  &  9.27 $\pm$ 6.01 \\ 
\hline
\textbf{DMV (ours)} & \textbf{91.60} $\pm$ \textbf{3.99} & \textbf{90.36} $\pm$ \textbf{3.87}  & \textbf{96.02} $\pm$ \textbf{3.20}
 &  \textbf{4.35} $\pm$ \textbf{2.82} & \textbf{3.84} $\pm$ \textbf{2.68} & \textbf{3.64} $\pm$ \textbf{2.93}  \\ 
\hline
\multirow{2}{*}{\textbf{Method}} & \multicolumn{2}{c|}{Dice Score \mbox{$\uparrow$[$\%$]}} &  \multicolumn{2}{c|}{Chamfer weighted \mbox{$\downarrow$}} & \multicolumn{2}{c|}{All predictions} \\ 
 & \multicolumn{2}{c|}{HI, MRI} &  \multicolumn{2}{c|}{HI, MRI} &  \multicolumn{2}{c|}{watertight} \\ 
\hline
Voxel2Mesh~\cite{wickramasinghe2020voxel2mesh} &  \multicolumn{2}{c|}{85.30 $\pm$ 3.20} & \multicolumn{2}{c|}{(1.3 $\pm$ 0.3) $\times$ $10^{-3}$} & \multicolumn{2}{c|}{No} \\ 
\hline
\textbf{DMV (ours)} &  \multicolumn{2}{c|}{\textbf{89.23} $\pm$ \textbf{3.08}} &  \multicolumn{2}{c|}{(\textbf{1.0} $\pm$ \textbf{0.1}) $\times$ $\textbf{10}^{-3}$} & \multicolumn{2}{c|}{\textbf{Yes}} \\ 
\hline
\end{tabularx}
\end{table*}

 \begin{figure*}
    \centering
    \includegraphics[width=.93\linewidth, angle=0]{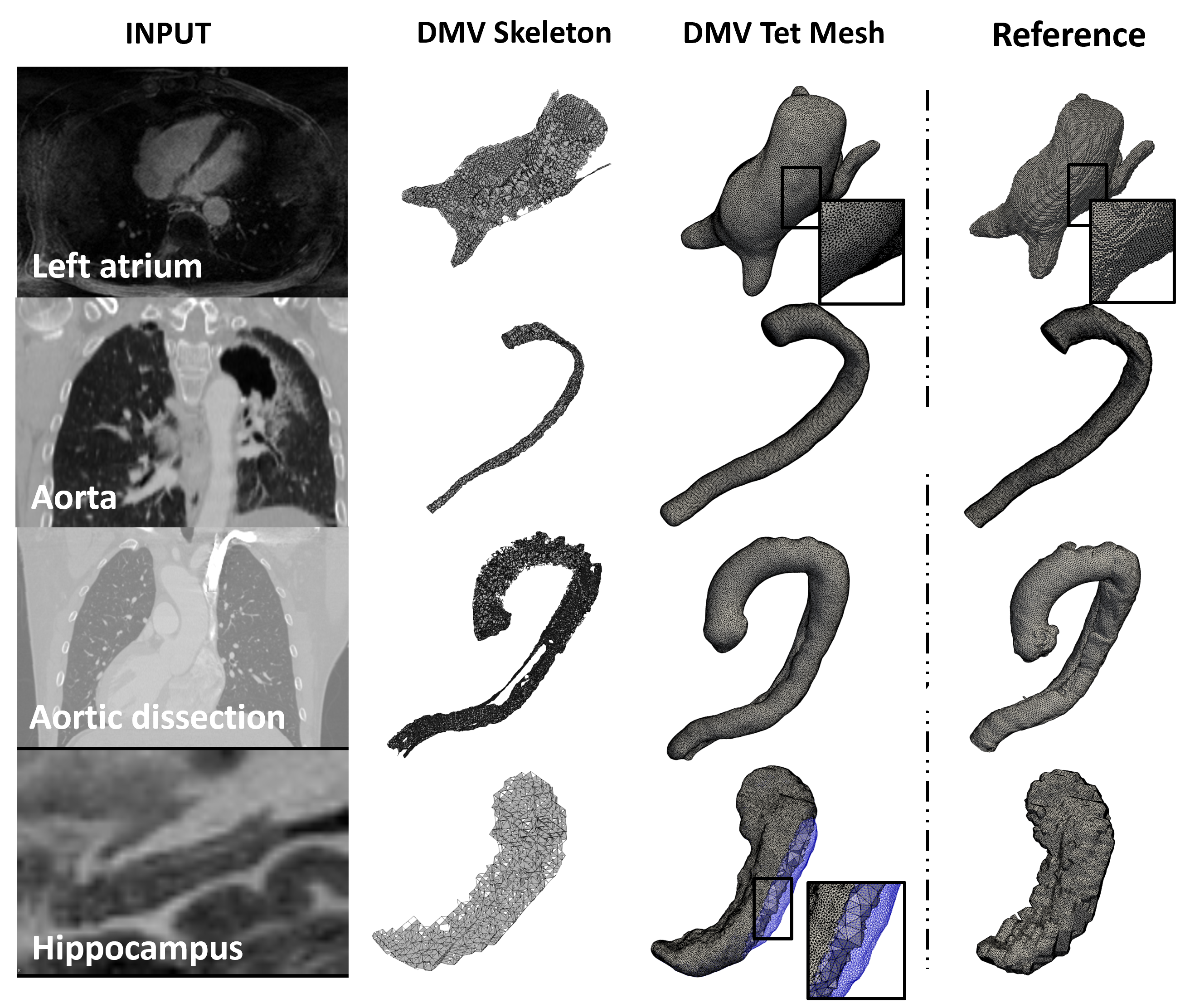}
    \caption{Qualitative results of test cases. From left to right: input section, predicted skeleton as a simplicial complex, tetrahedral mesh from the convolution surface of the skeleton, post-processed marching cubes triangulation as a test reference. The blue section cut of the tetrahedral mesh shows the inner volume.}
    \label{fig:2}
\end{figure*}

\subsection{Datasets} 

We tested our method on labeled public datasets of 154 left cardiac atria (MRI, \emph{2018 Left Atrium Segmentation Challenge}~\cite{xiong2021global}), 270 hippocampi (MRI, \emph{Medical Segmentation Decathlon}~\cite{simpson2019large}), 40 aortae (CT, \emph{SegThor Challenge}~\cite{lambert2020segthor}), and 100 aortic dissections (CT, \emph{Yao et al.}~\cite{yao2021imagetbad}). All datasets allow usage for research purposes. Volumes were resampled to an isotropic resolution of $1$~mm and randomly augmented during training with rotation and cropping. 

\subsection{Ablation study and quantitative analysis}

We performed an ablation study of the components of the loss function (\autoref{tab2}) on the left atria dataset. We used five-fold cross-validation, learning rate $l_r = 10^{-2}$, patch size $112\times112\times80$, batch size 8 and stochastic gradient descent. V-Net~\cite{milletari2016v} was used as a backbone, as it previously showed good performance in learning implicit representations~\cite{li2020shape}. Ablation showed a clear contribution of both Laplacian and MAT information to the loss function. Regularization of the UDF further increased the prediction quality. This result shows how joint topology and gradient information provide a more robust training than explicit shape matching only. With this final configuration, we tested DMV against the original backbone (V-Net) and the two most relevant methods based on implicit models (DDT~\cite{wang2020deep}, SASSNet~\cite{li2020shape}) in different modalities (\autoref{tab3}). Furthermore, we compared DMV with Voxel2Mesh~\cite{wickramasinghe2020voxel2mesh}, which directly generates surface meshes for the hippocampus. While all methods performed sufficiently well, DMV quantitatively outperformed all methods in these scenarios. Furthermore, DMV guarantees the generation of smooth and watertight surfaces thanks to the properties of convolution surfaces. 

\begin{table*}
    \centering
    \setlength{\tabcolsep}{0.5\tabcolsep}
        \caption{Qualitative evaluation of reconstruction quality. The study involved 12 field experts with multiple years of experience in cardiovascular modeling and simulation. Legend: 1 - Completely disagree; 2 - Somewhat disagree; 3 - Neutral; 4 - Somewhat agree; 5 - Completely agree.}
\begin{tabularx}{\linewidth}{| l |X|X|X|X|X|X|X|X|}
\hline
         \textbf{Question} & \multicolumn{5}{c|}{\textbf{Answer}} &\textbf{Median} & \textbf{Inter-quartile range}  \\
         & \textbf{1} & \textbf{2}  & \textbf{3} & \textbf{4}  & \textbf{5} & &  \\
         \hline
         Small structures are well preserved. & 0 & 0 & \cellcolor{gray!10}2  &  \cellcolor{gray!30}6 & \cellcolor{gray!20}4 & \cellcolor{gray!20}4 & \cellcolor{gray!5}1 \\
         \hline
         Artifacts are absent or negligible. & 0 & \cellcolor{gray!5}1 & \cellcolor{gray!5}1 &  \cellcolor{gray!10}4 & \cellcolor{gray!30}6 & \cellcolor{gray!25}4.5 & \cellcolor{gray!5}1 \\
         \hline
         Topology is well preserved. & 0 & 0 & 0  &  \cellcolor{gray!25}5 & \cellcolor{gray!30}6 & \cellcolor{gray!25}5 & \cellcolor{gray!5}1 \\
         \hline
         The mesh could easily be used in CFD simulations. & 0 & 0 & \cellcolor{gray!5}1  &  \cellcolor{gray!20}4 & \cellcolor{gray!35}7 & \cellcolor{gray!25}5 & \cellcolor{gray!5}1 \\
         \hline
         The cluster-based parametrization improves visualization. & 0 & 0 & 0  &  \cellcolor{gray!25}5 & \cellcolor{gray!35}7 & \cellcolor{gray!25}5 & \cellcolor{gray!5}1 \\
         \hline
         The cluster-based parametrization is helpful in CFD. & 0 & 0 & \cellcolor{gray!5}1  & \cellcolor{gray!20} 4 & \cellcolor{gray!35}7 & \cellcolor{gray!25}5 & \cellcolor{gray!5} 1 \\
\hline
         \textbf{Participants} & \multicolumn{5}{c|}{\textbf{Years}} & \textbf{Median} & \textbf{Inter-quartile range} \\
         & 1-2 & 3-4 & 5-6 & 7-8 & 9+ & &\\
         \hline
         Years of work experience in the field. & \cellcolor{gray!10}2 & \cellcolor{gray!10}2 & \cellcolor{gray!30}6 & 0 & \cellcolor{gray!5}1 & \cellcolor{gray!25}5 & \cellcolor{gray!5}1 \\
         \hline

         & \multicolumn{7}{c|}{\textbf{Job level}} \\
         & \multicolumn{2}{c|}{Lead scientist} & \multicolumn{2}{c|}{Senior research scientist} & \multicolumn{2}{c|}{Research scientist} & Research intern \\
         Current job level. & \multicolumn{2}{c|}{\cellcolor{gray!5}1} & \multicolumn{2}{c|}{\cellcolor{gray!25}5} & \multicolumn{2}{c|}{\cellcolor{gray!25}5} & \multicolumn{1}{c|}{\cellcolor{gray!5}1} \\
         \hline
         \hline
    \end{tabularx}
    \label{tab:questionnaire}
\end{table*}

\subsection{Qualitative analysis}

For visual understanding, \autoref{fig:2} provides a qualitative overview of the DMV output as skeleton and the reconstructed tetrahedral meshes. Example outputs are compared with the corresponding ground truth. It can be seen that the ground truth presents a higher number of discretization artifacts. Furthermore, a Likert scale questionnaire administered to senior biomedical research scientists showed great acceptance of the shapes generated for simulation, such as computational fluid dynamics (CFD) and visualization. The summary is provided in \autoref{tab:questionnaire}. Seven of the $12$ scientists ($58.3\%$) had at least five years of experience and six ($50.0\%$) had a senior or lead research position in industry or academia. All participants provided a neutral to extremely positive rank for the mesh outputs with the exception of one participant, who could still notice minor artifacts in some cases. 

\begin{figure*}[t]
\begin{minipage}{.5\textwidth}
    \centering
    \includegraphics[width=\linewidth]{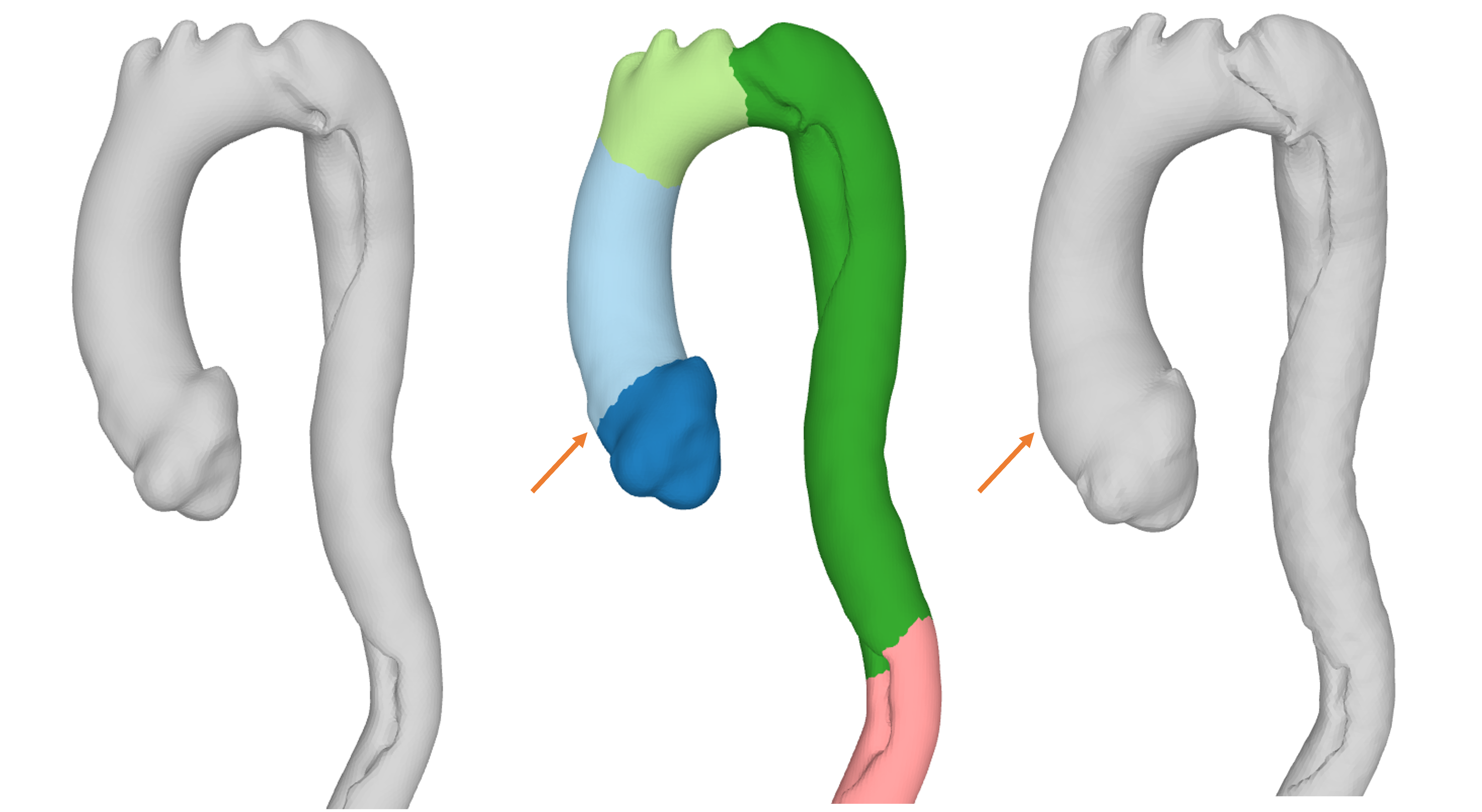}
    \caption{The co-segmentation has been used to modify the radii associated with the MAT points of the aortic root and the ascending aorta. As a result, it is possible to simulate the onset of pathological processes like aneurysmatic enlargements.}
    \label{fig:param}
\end{minipage}%
\vspace{0.03\textwidth}
\begin{minipage}{.5\textwidth}
    \centering
    \includegraphics[width=\linewidth]{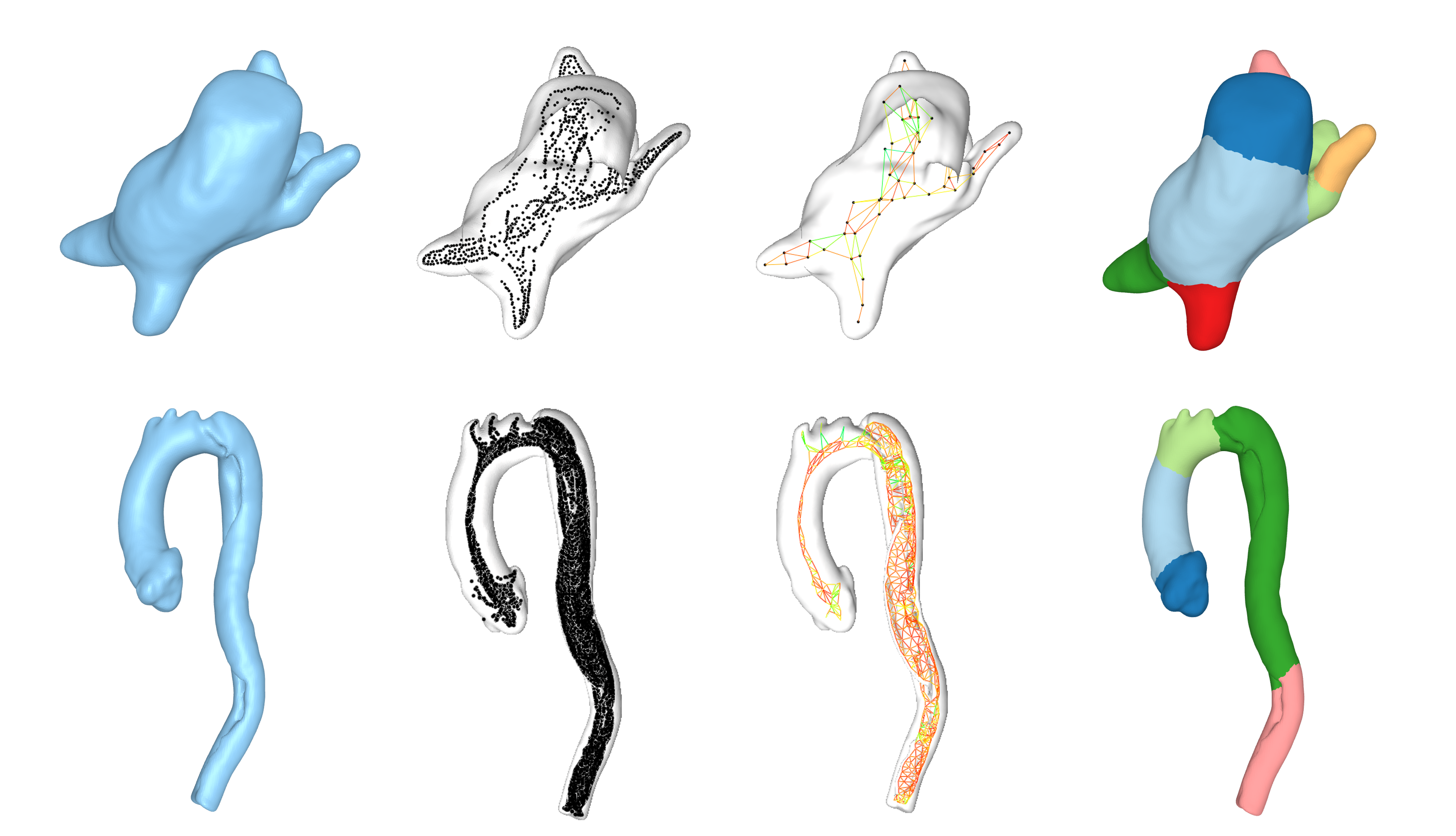}
    \caption{Examples of shape co-segmentation from the Q-MAT simplification~\cite{li2015q} of the predicted medial skeletons for a left cardiac atrium (top) and a dissected aorta (bottom). From the left, the reconstructed surface, the medial vertices, the simplified medial skeleton and the co-segmented surface. Without prior information, the co-segmentation correctly splits the shape into meaningful regions. The dissected aorta is split into aortic root, ascending aorta, aortic arch, descending aorta, and abdominal aorta.}
    \label{fig:cosegmentation}
\end{minipage}

\end{figure*}

\subsection{Co-segmentation and parameterization}

A particularity of the simplicial complex is that it contains topological information and can therefore be used for shape co-segmentation~\cite{lin2020seg}. In \autoref{fig:cosegmentation}, we provide examples of how our simplified skeletons can be combined with SEG-MAT~\cite{lin2020seg} for unsupervised shape co-segmentation~\cite{lin2020seg}, without the need for any additional post-processing. We also provide an example of cluster-based shape parameterization (\autoref{fig:param}), where the co-segmented cluster of the aortic root (blue) has been deformed using a linear parametrization along the cluster. The resulting effect simulates the presence of an aneurysmatic dilatation. Such parametrizations are often non-trivial~\cite{zhang2019real}, but relevant in sensitivity analysis of biomechanics simulations~\cite{garcia2018sensitivity}. The dilation effect was obtained by multiplying the radii of the MAT points by a dilation coefficient modulated by a Gaussian distribution centered on the junction between the blue (root) and the light blue (ascending aorta) clusters. The modulation of the MAT radii allows one to recompute a new convolution surface and generate an updated mesh that is still smooth and watertight.  

\begin{figure*}[!t]
    \centering
    \includegraphics[width=\linewidth]{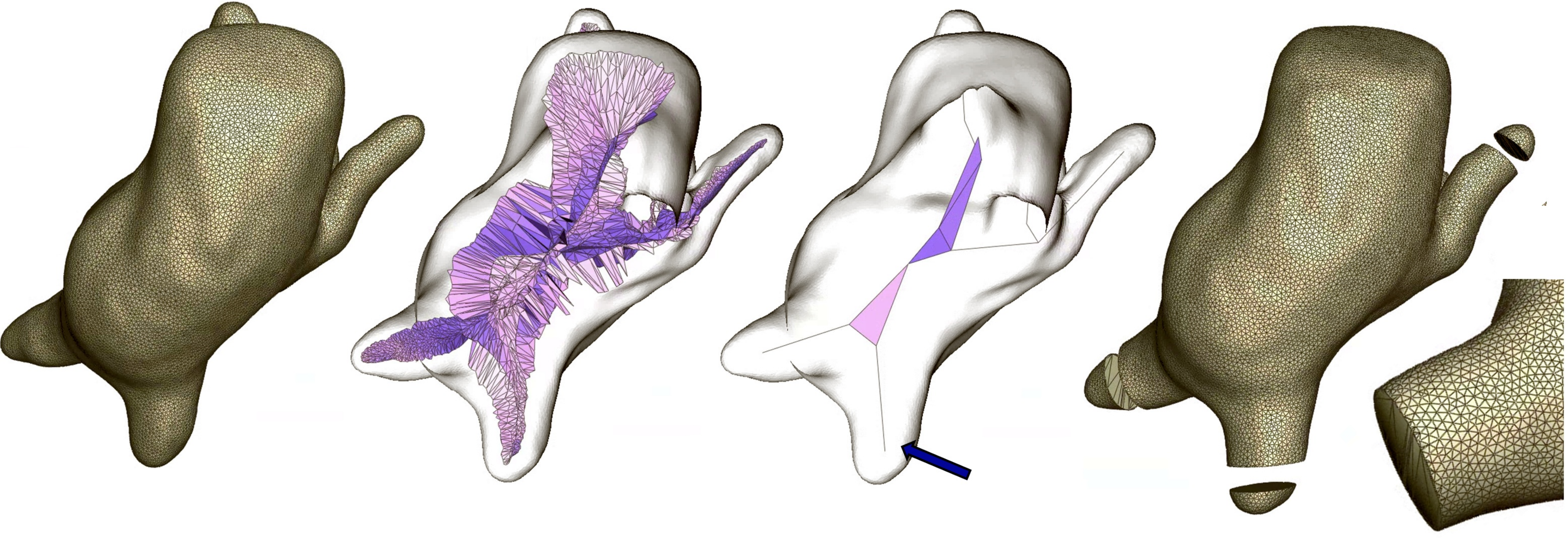}
    \caption{From left to right: Predicted DMV mesh of a left atrium, MAT represented as simplicial complex, Q-MAT simplification where the extremities are described with segments, truncated mesh where inflow and outflow interfaces are modeled as flat cuts, rotated view showing additional interface. }
    \label{fig:interfaces_mesh}
\end{figure*}

\subsection{Definition of inflow and outflow interfaces}

For computational fluid dynamics (CFD), the generated meshes require the definition of inflow and outflow interfaces. Manual definition of such interfaces can be time-consuming. A common alternative is to place them at the extremities (leaves) of the shape centerline and orient them orthogonally to the centerline itself. The predicted DMV skeletons are generally too complex to quickly determine such locations. However, the interface points are mostly tubular, and oversimplifying the MAT leads to a simplicial complex where the leaf elements are depicted as simple segments, allowing quick determination of the leaf elements and their orientations (\autoref{fig:interfaces_mesh}).

\subsection{Numerical simulation}

\begin{figure}
    \centering
    \includegraphics[width=.48\linewidth]{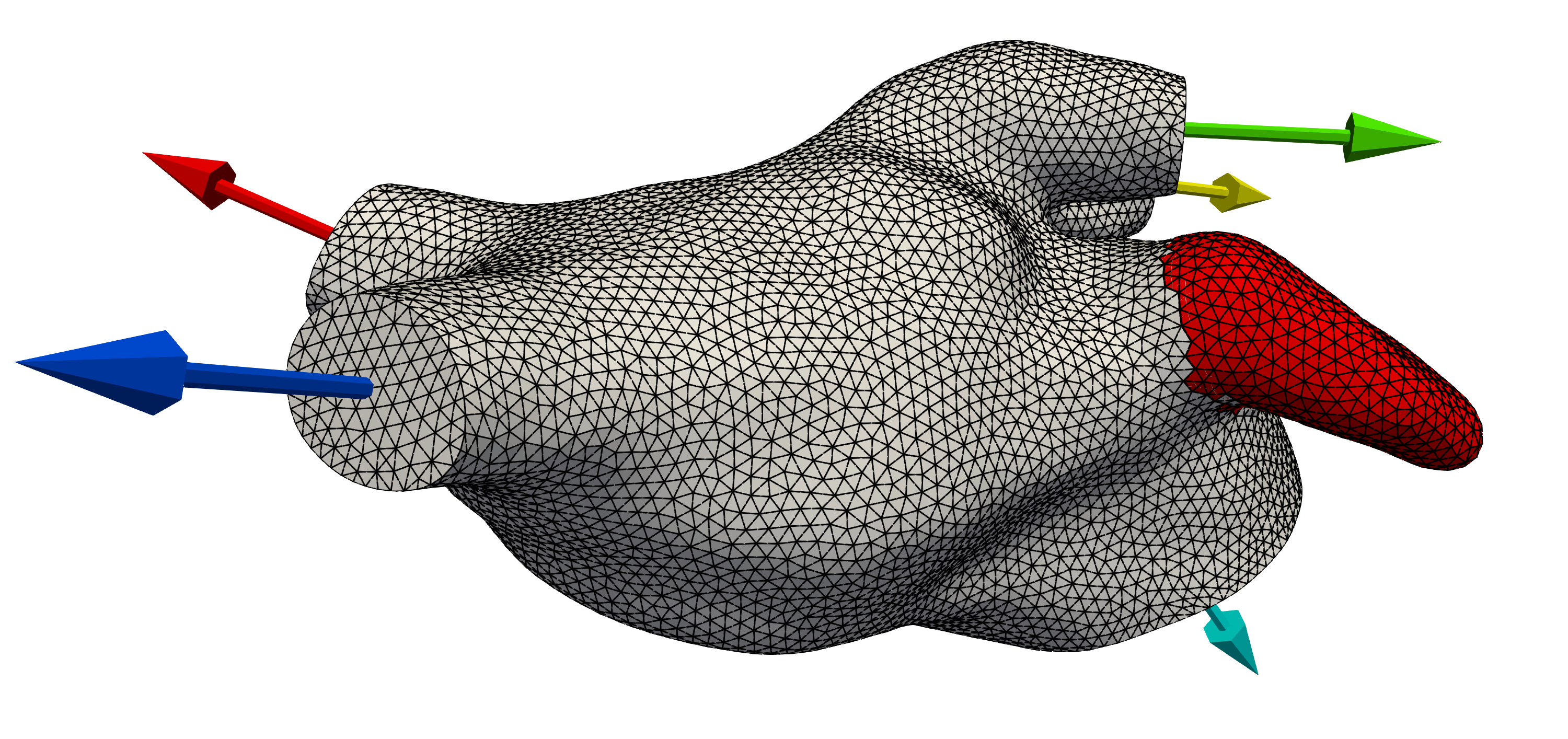}
    \hfill
    \includegraphics[width=.48\linewidth]{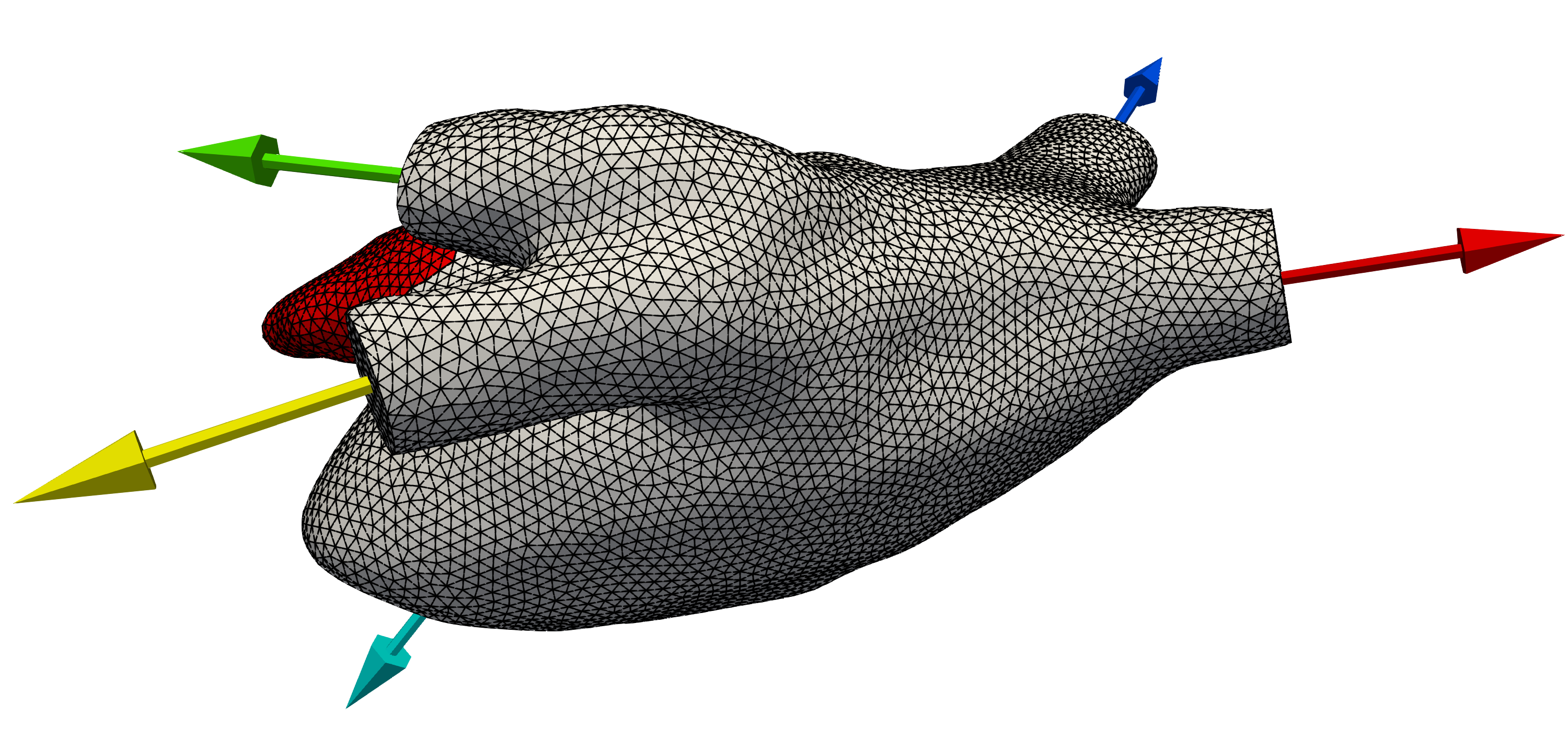}
    \caption{Tetrahedral grid employed for fluid mechanics simulation: anterior (top) and posterior view (bottom) with the left atrium appendage highlighted in red and outward normals on the LSPV (green), LIPV (yellow), RSPV (blue), RIPV (red) and MV (cyan).}
    \label{fig:num_grid}
\end{figure}

Patient-specific simulations of the cardiovascular system and its pathologies are of increasing importance in clinical practice, as they allow virtual surgery, facilitate enhanced planning or can be employed in medical device design. In this context, the construction of suitable (volumetric) meshes is a major challenge. Mesh quality significantly affects the time to find a solution and thereby influences the clinical relevance of computational methods in medicine. For example, the triangular surface representations derived via the proposed framework can be readily employed for biomedical applications such as fluid and solid mechanics or fluid--structure interaction like those shown in Bäumler et al.~\cite{baumler2020fluid} and Schussnig et al.~\cite{Schussnig2021b,Schussnig2022c, Schussnig2021d}. Here, the blood flow in the left atrium is simulated. For this purpose, a tetrahedral grid is constructed using \texttt{Gmsh}~\cite{Geuzaine2009Gmsh} based on a triangular surface mesh generated by DMV (\autoref{fig:num_grid}). To show the suitability of DMV meshes for numerical simulations, we consider a stationary atrium, demonstrating suitability of the grids for CFD. Identical volumetric flow rates are prescribed in the left and right inferior and superior pulmonary veins (PV) together with a zero reference pressure at the mitral valve (MV), following Dueñas-Pamplona et al.~\cite{DuenasPamplona2021atriumCFD}.

Blood velocity $\mathbf{u}$ and pressure $p$ in the domain $\Omega$ are governed by the Navier--Stokes equations for incompressible flow of a generalized Newtonian fluid,
\begin{align}
    \rho \frac{\partial}{\partial t}\mathbf{u} 
    + 
    \rho \mathbf{u} \cdot \nabla \mathbf{u}
    - 
    2 \nabla \cdot \left(\mu \nabla^\mathrm{S}\mathbf{u} \right)
    + 
    \nabla p
    &= 
    \mathbf{0}
    & \text{in }\Omega
    ,
    \\
    \nabla \cdot \mathbf{u} 
    &= 
    0
    & \text{in }\Omega
    ,
\end{align}
where $\rho = 1060~\text{kg/m}^3$ is the density; $\mu$, the viscosity, and $\nabla^S$, the symmetric gradient. The viscosity is given by
\begin{equation}
    \label{eqn:fluid_model}
    \mu (\dot\gamma) = \mu_\infty + (\mu_0 - \mu_\infty) \left( 1.0 + (\lambda \dot\gamma)^2 \right) ^{\frac{n-1}{2}}
    ,
\end{equation}
with the shear rate $\dot{\gamma} = \sqrt{2 \nabla^\mathrm{S} \mathbf{u} : \nabla^\mathrm{S} \mathbf{u}}$. The physiological parameters $\mu_\infty=3.22$~mPas, $\mu_0=27.21$~mPas, $\lambda=1.556$~1/s and $n=0.462$ follow the settings of Ranftl et al.~\cite{Ranftl2021}. Simulations are carried out in \texttt{ExaDG}~\cite{ExaDG2020}, a multi-purpose, higher-order discontinuous Galerkin flow solver based on the \texttt{deal.II} finite element library~\cite{dealII95}, employing Taylor--Hood finite elements and hybrid multigrid methods~\cite{Fehn2018,Fehn2020}.

\begin{figure*}
    \centering
        \begin{overpic}[width=0.31\textwidth,draft=false]{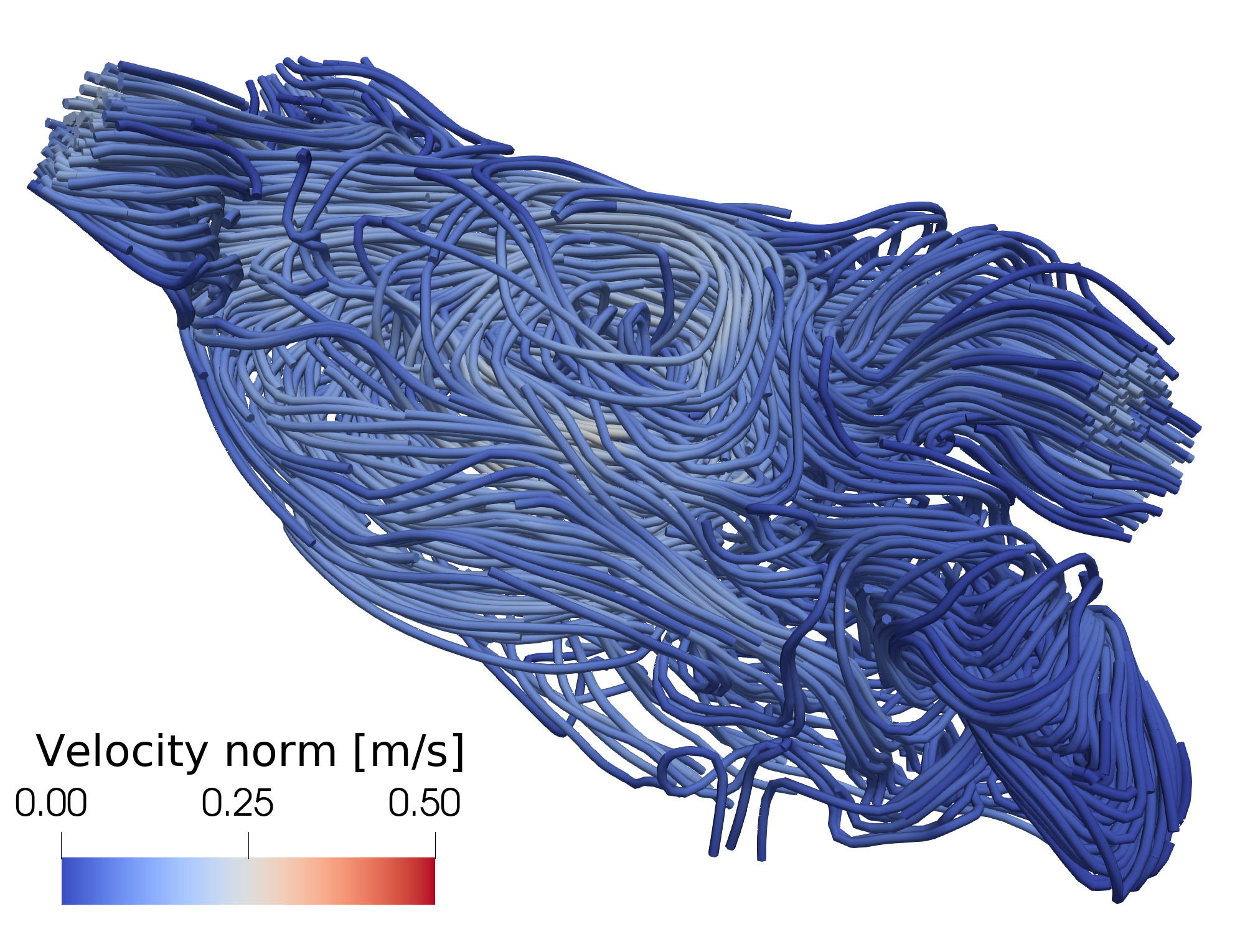}
        \end{overpic}
        \begin{overpic}[width=0.31\textwidth,draft=false]{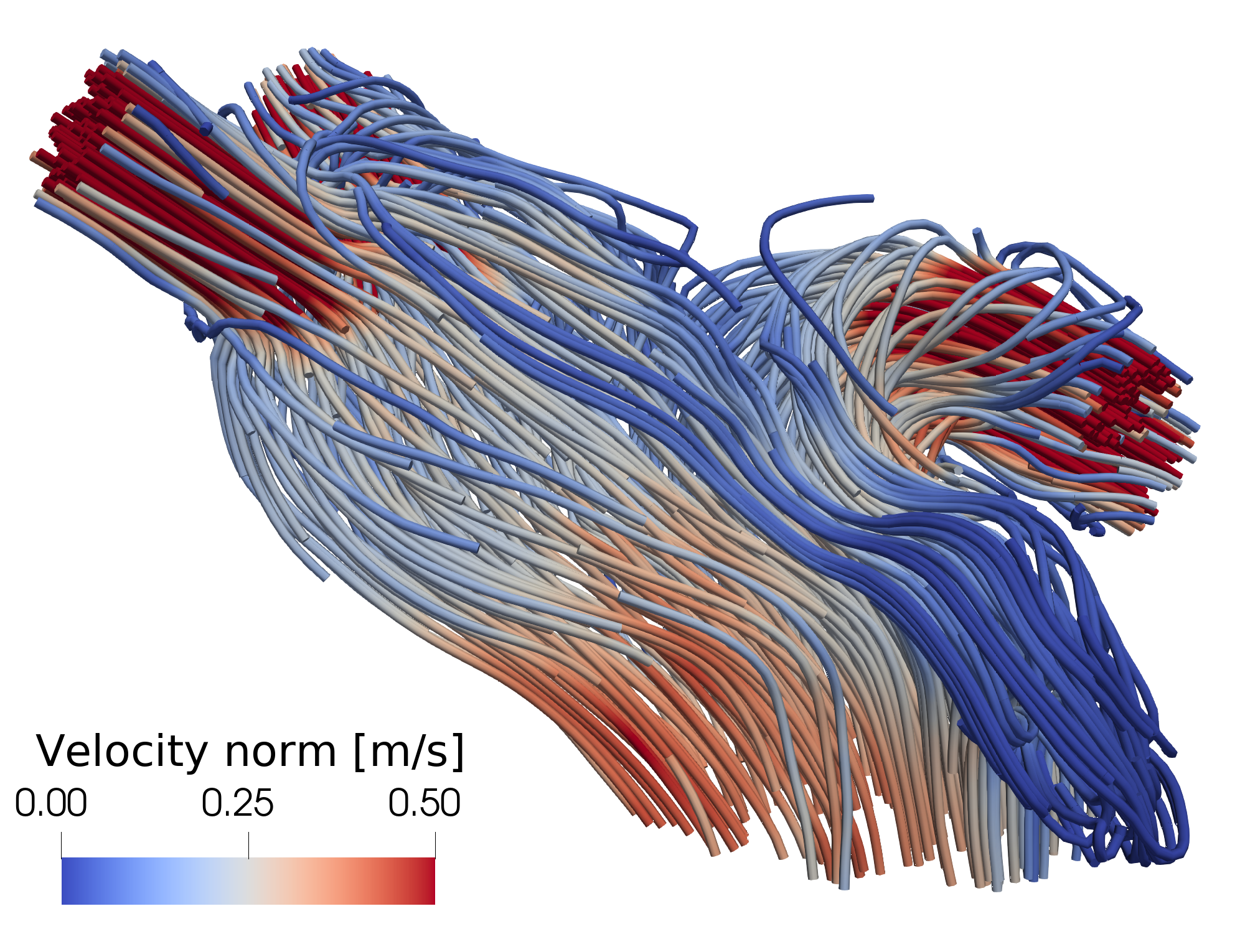}
        \end{overpic}
        \begin{overpic}[width=0.31\textwidth,draft=false]{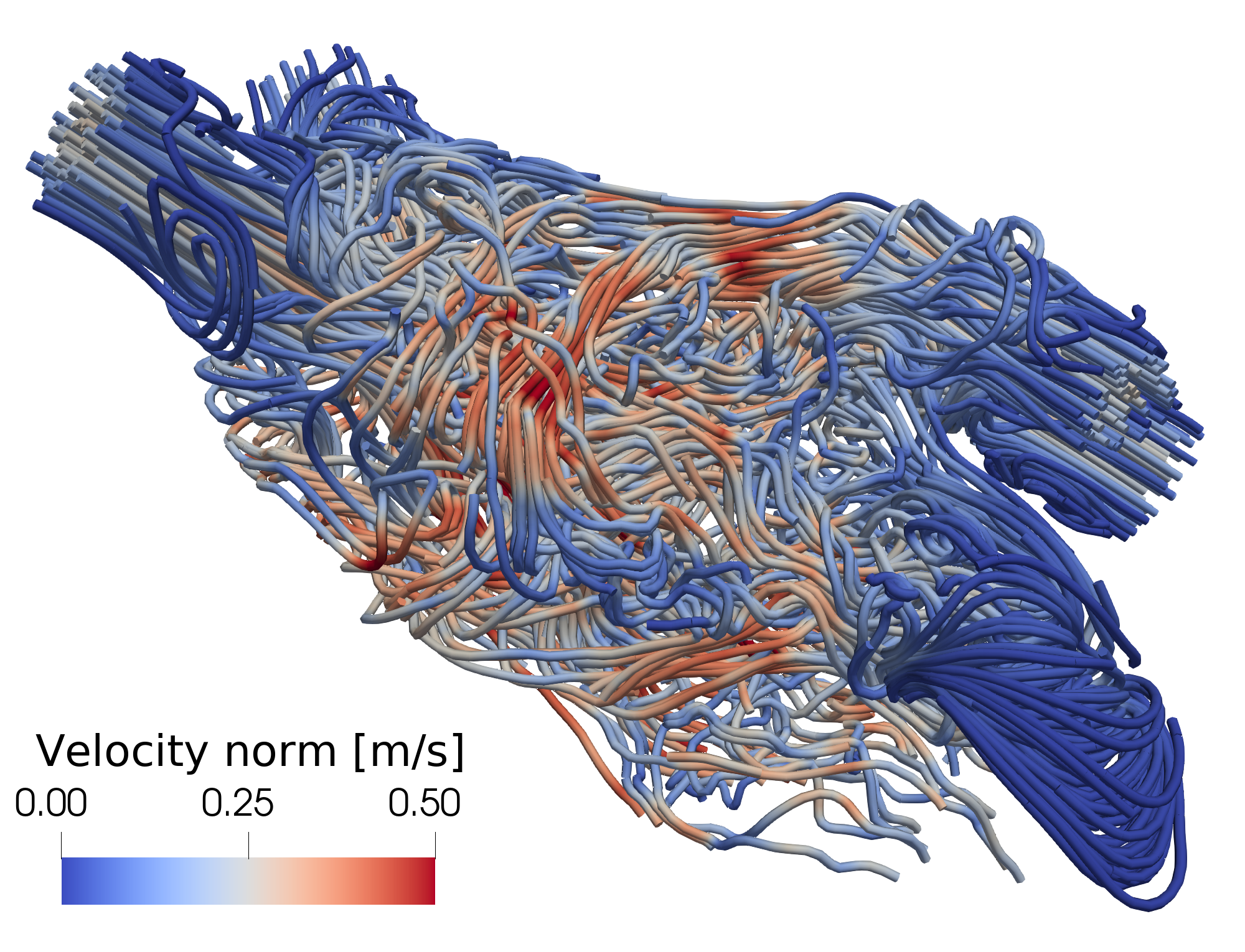}
        \end{overpic}
        \\
        \begin{overpic}[width=0.31\textwidth,draft=false]{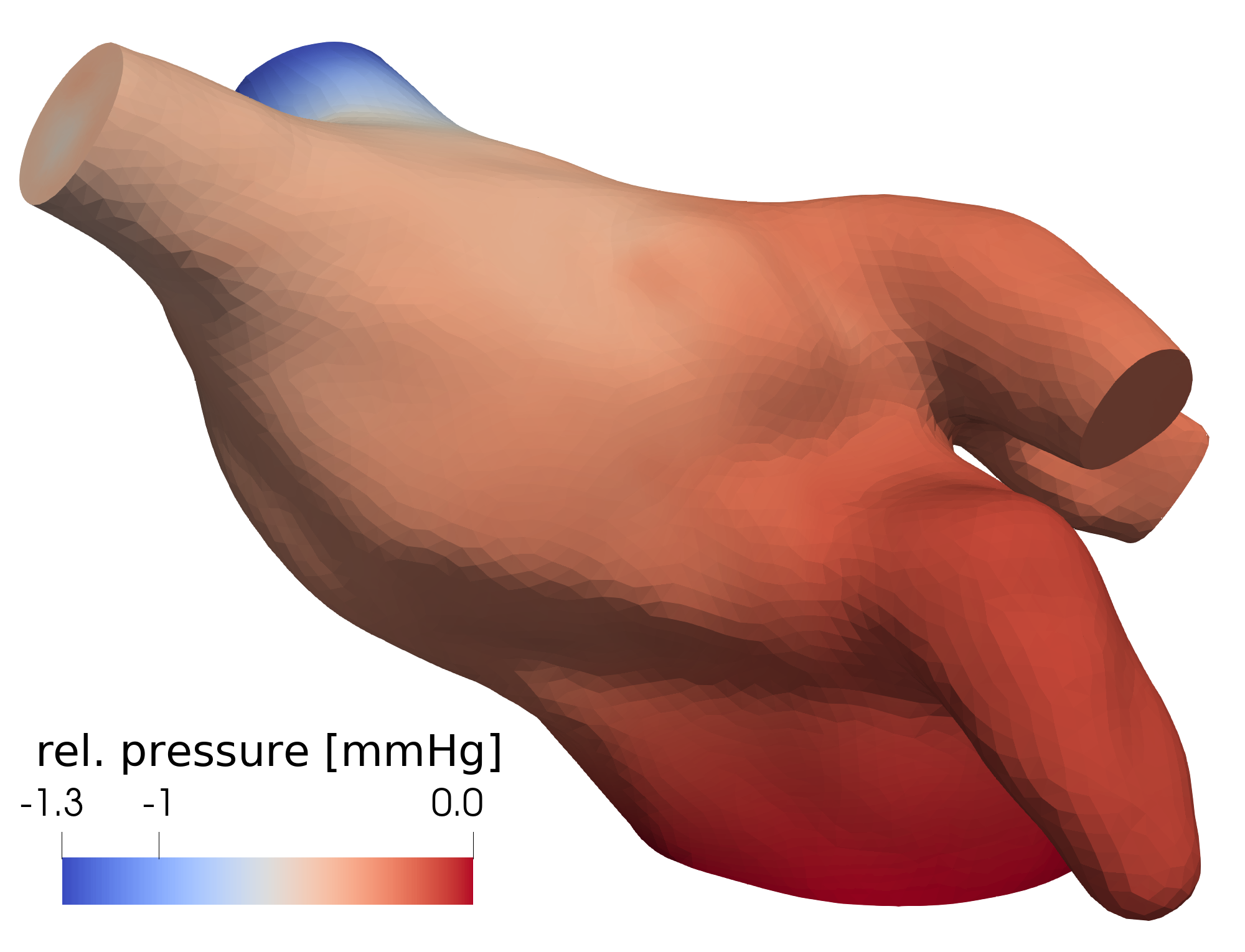}
           \put(45,65){\frame{\includegraphics[width=0.065\textwidth]{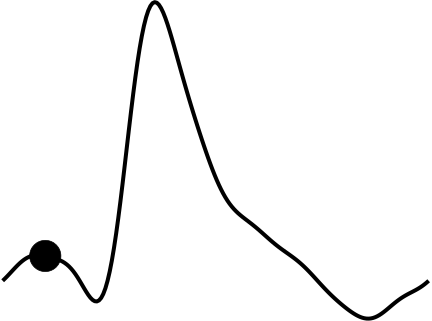}}}
        \end{overpic}
        \begin{overpic}[width=0.31\textwidth,draft=false]{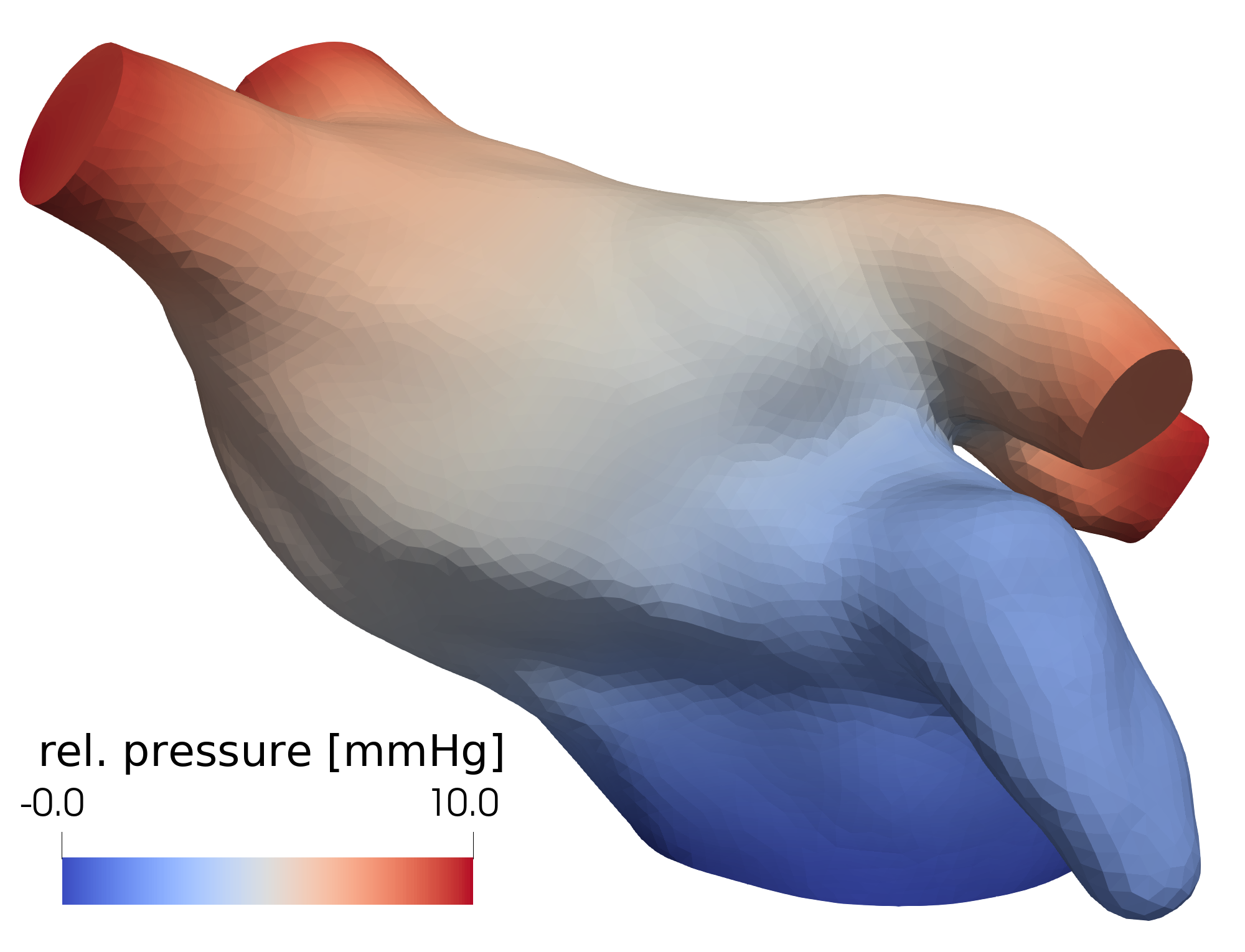}
            \put(45,65){\frame{\includegraphics[width=0.065\textwidth]{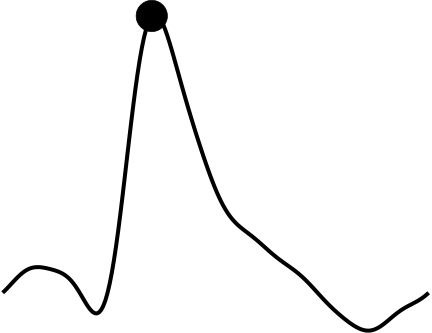}}}
        \end{overpic}
        \begin{overpic}[width=0.31\textwidth,draft=false]{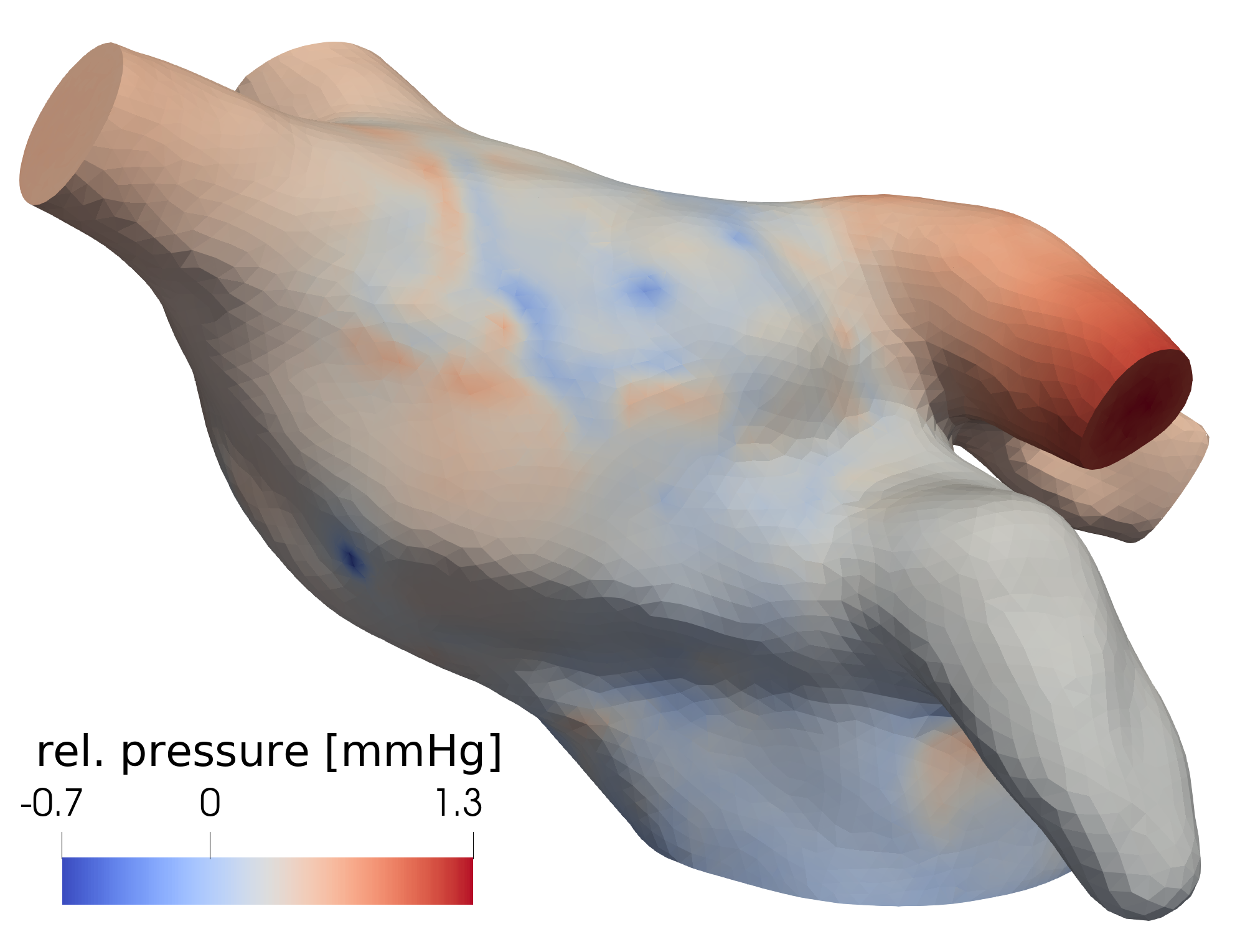}
            \put(45,65){\frame{\includegraphics[width=0.065\textwidth]{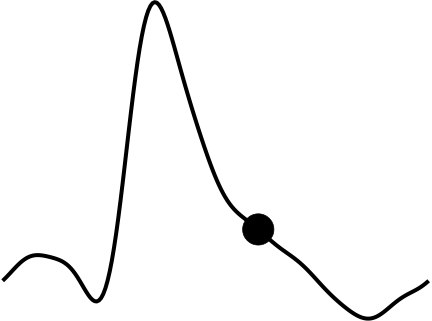}}}
        \end{overpic}
    \caption{Streamlines colored by velocity norm (top row) and relative pressure (bottom row) in diastole (left column), peak systole (middle column) and late systole (right column) as indicated by the inflow profile scale. Streamlines show a largely laminar flow, which is disorganized in diastole and stagnant in the left atrium appendage, while the pressure gradient from PV to MV reaches 10 mmHg in systole.}
    \label{fig:num_results}
\end{figure*}

Snapshots of the flow field and pressure distribution at three distinct points (diastole, peak systole and late systole) in the fourth cardiac cycle considered show mainly laminar flow from the PV to the MV. During diastole, the flow becomes disorganized, and the expected vortical structures are observed (\autoref{fig:num_results}). Flow in the left atrium appendage is low, i.e., almost stagnant throughout the cardiac cycle. Due to the regular grid, iteration counts in the linear solvers stay low throughout the entire simulation, while the approximation quality seems adequate using elements with a target edge length of $0.8$--$1.8~$mm and second-order velocity interpolation. Further aspects such as a detailed analysis of the fluid dynamics, a grid convergence study and model verification are omitted for brevity. However, decent grid quality is required for CFD and is therefore implicitly demonstrated in this example.

\section{Conclusion}

We presented an automatic strategy for adaptive and parametric modeling of anatomical shapes from raw 3D imaging data. We showed how an approximation of the medial axis transform can be generated through regularized voxel-to-voxel transformations. The presented strategy is the first to address the problem of topology preservation for adaptive geometry reconstruction from learned 3D discrete implicit fields of different organs and imaging modalities. Additionally, we showed how this representation could provide a versatile medium for various tasks ranging from volumetric meshing to shape co-segmentation. A qualitative expert evaluation based on a Likert scale questionnaire showed a positive perception of the generated shapes. Additionally, a CFD simulation confirmed the direct applicability of the DMV output in FEM-based numerical simulations. 
In future work, we will further examine the capabilities of DMV for procedural and neural shape augmentation and the possibility of relying on DMV for the generation of structured hexahedral meshes for different shapes~\cite{bovsnjak2023higher,viville2023meso} and their application in in-silico trials~\cite{sarrami2021silico}. 

\bibliographystyle{ieeetr}
\bibliography{references}

\begin{thebibliography}{10}

\bibitem{mistelbauer2021implicit}
G.~Mistelbauer, C.~R{\"o}ssl, K.~B{\"a}umler, B.~Preim, and D.~Fleischmann, ``Implicit modeling of patient-specific aortic dissections with elliptic fourier descriptors,'' in {\em Comput Graph Forum}, vol.~40, pp.~423--34, 2021.

\bibitem{oeltze2005visualization}
S.~Oeltze and B.~Preim, ``Visualization of vasculature with convolution surfaces: method, validation and evaluation,'' {\em IEEE Trans Med Imaging}, vol.~24, no.~4, pp.~540--48, 2005.

\bibitem{li2021autoimplant}
J.~Li {\em et~al.}, ``Autoimplant 2020-first miccai challenge on automatic cranial implant design,'' {\em IEEE Trans Med Imaging}, vol.~40, no.~9, pp.~2329--42, 2021.

\bibitem{baumler2020fluid}
K.~B{\"a}umler {\em et~al.}, ``Fluid--structure interaction simulations of patient-specific aortic dissection,'' {\em Biomechanics and Modeling in Mechanobiology}, vol.~19, no.~5, pp.~1607--28, 2020.

\bibitem{shad2021patient}
R.~Shad {\em et~al.}, ``Patient-specific computational fluid dynamics reveal localized flow patterns predictive of post--left ventricular assist device aortic incompetence,'' {\em Circulation: Heart Failure}, vol.~14, no.~7, p.~e008034, 2021.

\bibitem{antiga2008image}
L.~Antiga, M.~Piccinelli, L.~Botti, B.~Ene-Iordache, A.~Remuzzi, and D.~A. Steinman, ``An image-based modeling framework for patient-specific computational hemodynamics,'' {\em Medical \& biological engineering \& computing}, vol.~46, pp.~1097--1112, 2008.

\bibitem{kerrien2017blood}
E.~Kerrien, A.~Yureidini, J.~Dequidt, C.~Duriez, R.~Anxionnat, and S.~Cotin, ``Blood vessel modeling for interactive simulation of interventional neuroradiology procedures,'' {\em Med Image Anal}, vol.~35, pp.~685--98, 2017.

\bibitem{bovsnjak2023higher}
D.~Bo{\v{s}}njak, A.~Pepe, R.~Schussnig, D.~Schmalstieg, and T.-P. Fries, ``Higher-order block-structured hex meshing of tubular structures,'' {\em Engineering with Computers}, pp.~1--21, 2023.

\bibitem{burris2022vascular}
N.~S. Burris {\em et~al.}, ``Vascular deformation mapping for {CT} surveillance of thoracic aortic aneurysm growth,'' {\em Radiology}, vol.~302, no.~1, pp.~218--25, 2022.

\bibitem{sarrami2021silico}
A.~Sarrami-Foroushani {\em et~al.}, ``In-silico trial of intracranial flow diverters replicates and expands insights from conventional clinical trials,'' {\em Nature Communications}, vol.~12, no.~1, pp.~1--12, 2021.

\bibitem{wang2020deep}
Y.~Wang {\em et~al.}, ``Deep distance transform for tubular structure segmentation in {CT} scans,'' in {\em Proc. IEEE/CVF Conference on Computer Vision and Pattern Recognition}, pp.~3833--42, 2020.

\bibitem{li2020shape}
S.~Li, C.~Zhang, and X.~He, ``Shape-aware semi-supervised 3d semantic segmentation for medical images,'' in {\em International Conference on Medical Image Computing and Computer-Assisted Intervention}, pp.~552--61, Springer, 2020.

\bibitem{hu2019mat}
J.~Hu, B.~Wang, L.~Qian, Y.~Pan, X.~Guo, L.~Liu, and W.~Wang, ``Mat-net: Medial axis transform network for 3d object recognition.,'' in {\em IJCAI}, pp.~774--781, 2019.

\bibitem{lin2021point2skeleton}
C.~Lin, C.~Li, Y.~Liu, N.~Chen, Y.-K. Choi, and W.~Wang, ``Point2skeleton: Learning skeletal representations from point clouds,'' in {\em Proc. IEEE/CVF Conference on Computer Vision and Pattern Recognition}, pp.~4277--86, 2021.

\bibitem{suarez2019anisotropic}
A.~J.~F. Su{\'a}rez, E.~Hubert, and C.~Zanni, ``Anisotropic convolution surfaces,'' {\em Computers \& Graphics}, vol.~82, pp.~106--16, 2019.

\bibitem{newman2006survey}
T.~S. Newman and H.~Yi, ``A survey of the marching cubes algorithm,'' {\em Computers \& Graphics}, vol.~30, no.~5, pp.~854--79, 2006.

\bibitem{hong2019high}
Q.~Hong, Q.~Li, B.~Wang, K.~Liu, and Q.~Qi, ``High precision implicit modeling for patient-specific coronary arteries,'' {\em IEEE Access}, vol.~7, pp.~72020--29, 2019.

\bibitem{piccinelli2009framework}
M.~Piccinelli, A.~Veneziani, D.~A. Steinman, A.~Remuzzi, and L.~Antiga, ``A framework for geometric analysis of vascular structures: Application to cerebral aneurysms,'' {\em IEEE Transactions on Medical Imaging}, vol.~28, no.~8, pp.~1141--1155, 2009.

\bibitem{wickramasinghe2020voxel2mesh}
U.~Wickramasinghe, E.~Remelli, G.~Knott, and P.~Fua, ``Voxel2mesh: 3d mesh model generation from volumetric data,'' in {\em International Conference on Medical Image Computing and Computer-Assisted Intervention}, pp.~299--308, 2020.

\bibitem{pak2021distortion}
D.~H. Pak {\em et~al.}, ``Distortion energy for deep learning-based volumetric finite element mesh generation for aortic valves,'' in {\em International Conference on Medical Image Computing and Computer-Assisted Intervention}, pp.~485--94, Springer, 2021.

\bibitem{kanamori2008gpu}
Y.~Kanamori, Z.~Szego, and T.~Nishita, ``Gpu-based fast ray casting for a large number of metaballs,'' {\em Comput Graph Forum}, vol.~27, no.~2, pp.~351--60, 2008.

\bibitem{fries2017higher}
T.~Fries, S.~Omerovi{\'c}, D.~Sch{\"o}llhammer, and J.~Steidl, ``Higher-order meshing of implicit geometries, {Part I}:{I}ntegration and interpolation in cut elements,'' {\em Comput Methods Appl Mech Eng}, vol.~313, pp.~759--84, 2017.

\bibitem{torres2020simplicial}
J.~J. Torres and G.~Bianconi, ``Simplicial complexes: higher-order spectral dimension and dynamics,'' {\em J. phys. Complex}, vol.~1, no.~1, p.~015002, 2020.

\bibitem{edelsbrunner2022computational}
H.~Edelsbrunner and J.~L. Harer, {\em Computational topology: an introduction}.
\newblock American Mathematical Society, 2022.

\bibitem{giesen2009scale}
J.~Giesen, B.~Miklos, M.~Pauly, and C.~Wormser, ``The scale axis transform,'' in {\em Proc. Annual Symposium on Computational Geometry}, pp.~106--15, 2009.

\bibitem{li2015q}
P.~Li, B.~Wang, F.~Sun, X.~Guo, C.~Zhang, and W.~Wang, ``Q-mat: Computing medial axis transform by quadratic error minimization,'' {\em ACM Trans. Graph.}, vol.~35, no.~1, pp.~1--16, 2015.

\bibitem{dou2021coverage}
Z.~Dou {\em et~al.}, ``Coverage axis: Inner point selection for 3d shape skeletonization,'' {\em Comput Graph Forum}, vol.~41, no.~2, pp.~2329--42, 2022.

\bibitem{yan2018voxel}
Y.~Yan, D.~Letscher, and T.~Ju, ``Voxel cores: Efficient, robust, and provably good approximation of 3d medial axes,'' {\em ACM Trans. Graph.}, vol.~37, no.~4, pp.~1--13, 2018.

\bibitem{rebain2019lsmat}
D.~Rebain {\em et~al.}, ``{LSMAT} least squares medial axis transform,'' in {\em Comput Graph Forum}, vol.~38, pp.~5--18, 2019.

\bibitem{lin2020seg}
C.~Lin {\em et~al.}, ``Seg-mat: 3d shape segmentation using medial axis transform,'' {\em IEEE Trans Vis Comput Graph}, 2020.

\bibitem{park2019deepsdf}
J.~J. Park, P.~Florence, J.~Straub, R.~Newcombe, and S.~Lovegrove, ``Deep{SDF}: Learning continuous signed distance functions for shape representation,'' in {\em Proc. IEEE/CVF Conference on Computer Vision and Pattern Recognition}, pp.~165--74, 2019.

\bibitem{rebain2021deep}
D.~Rebain, K.~Li, V.~Sitzmann, S.~Yazdani, K.~M. Yi, and A.~Tagliasacchi, ``Deep medial fields,'' {\em arXiv preprint arXiv:2106.03804}, 2021.

\bibitem{genova2020local}
K.~Genova, F.~Cole, A.~Sud, A.~Sarna, and T.~Funkhouser, ``Local deep implicit functions for 3d shape,'' in {\em Proc. IEEE/CVF Conference on Computer Vision and Pattern Recognition}, pp.~4857--66, 2020.

\bibitem{yang2020p2mat}
B.~Yang {\em et~al.}, ``{P2MAT-Net}: Learning medial axis transform from sparse point clouds,'' {\em Comput Aided Geom Des}, vol.~80, p.~101874, 2020.

\bibitem{erler2020points2surf}
P.~Erler, P.~Guerrero, S.~Ohrhallinger, N.~J. Mitra, and M.~Wimmer, ``Points2surf: Learning implicit surfaces from point clouds,'' in {\em European Conference on Computer Vision}, pp.~108--124, 2020.

\bibitem{peng2020convolutional}
S.~Peng, M.~Niemeyer, L.~Mescheder, M.~Pollefeys, and A.~Geiger, ``Convolutional occupancy networks,'' in {\em European Conference on Computer Vision}, pp.~523--40, 2020.

\bibitem{zhao2022segmentation}
J.~Zhao, J.~Zhao, S.~Pang, and Q.~Feng, ``Segmentation of the true lumen of aorta dissection via morphology-constrained stepwise deep mesh regression,'' {\em IEEE Trans Med Imaging}, 2022.

\bibitem{ma2020distance}
J.~Ma {\em et~al.}, ``How distance transform maps boost segmentation {CNN}s: an empirical study,'' in {\em Medical Imaging with Deep Learning}, pp.~479--92, PMLR, 2020.

\bibitem{lin2023structure}
Z.~Lin, D.~Wei, A.~Gupta, X.~Liu, D.~Sun, and H.~Pfister, ``Structure-preserving instance segmentation via skeleton-aware distance transform,'' in {\em International Conference on Medical Image Computing and Computer-Assisted Intervention}, pp.~529--539, Springer, 2023.

\bibitem{zhang2023anatomy}
X.~Zhang, K.~Sun, D.~Wu, X.~Xiong, J.~Liu, L.~Yao, S.~Li, Y.~Wang, J.~Feng, and D.~Shen, ``An anatomy-and topology-preserving framework for coronary artery segmentation,'' {\em IEEE Transactions on Medical Imaging}, 2023.

\bibitem{xia2011fast}
H.~Xia and P.~G. Tucker, ``Fast equal and biased distance fields for medial axis transform with meshing in mind,'' {\em Applied Mathematical Modelling}, vol.~35, no.~12, pp.~5804--19, 2011.

\bibitem{liao2020real}
M.~Liao, Z.~Wan, C.~Yao, K.~Chen, and X.~Bai, ``Real-time scene text detection with differentiable binarization,'' in {\em Proc. AAAI conference on artificial intelligence}, vol.~34, pp.~11474--81, 2020.

\bibitem{shit2021cldice}
S.~Shit {\em et~al.}, ``cldice-a novel topology-preserving loss function for tubular structure segmentation,'' in {\em Proceedings of the IEEE/CVF Conference on Computer Vision and Pattern Recognition}, pp.~16560--16569, 2021.

\bibitem{mccormack1998creating}
J.~McCormack and A.~Sherstyuk, ``Creating and rendering convolution surfaces,'' in {\em Computer Graphics Forum}, vol.~17, pp.~113--20, 1998.

\bibitem{kazhdan2006poisson}
M.~Kazhdan, M.~Bolitho, and H.~Hoppe, ``Poisson surface reconstruction,'' in {\em Proc. EG Symposium on Geometry Processing}, vol.~7, 2006.

\bibitem{Geuzaine2009Gmsh}
C.~Geuzaine and J.~Remacle, ``Gmsh: A 3-d finite element mesh generator with built-in pre- and post-processing facilities,'' {\em Int J Numer Methods Eng}, vol.~79, no.~11, pp.~1309--1331, 2009.

\bibitem{milletari2016v}
F.~Milletari, N.~Navab, and S.-A. Ahmadi, ``V-net: Fully convolutional neural networks for volumetric medical image segmentation,'' in {\em International conference on 3D vision (3DV)}, pp.~565--71, 2016.

\bibitem{xiong2021global}
Z.~Xiong {\em et~al.}, ``A global benchmark of algorithms for segmenting the left atrium from late gadolinium-enhanced cardiac magnetic resonance imaging,'' {\em Medical Image Analysis}, vol.~67, p.~101832, 2021.

\bibitem{simpson2019large}
A.~L. Simpson {\em et~al.}, ``A large annotated medical image dataset for the development and evaluation of segmentation algorithms,'' {\em arXiv:1902.09063}, 2019.

\bibitem{lambert2020segthor}
Z.~Lambert, C.~Petitjean, B.~Dubray, and S.~Kuan, ``Segthor: Segmentation of thoracic organs at risk in ct images,'' in {\em Proc. IPTA}, pp.~1--6, IEEE, 2020.

\bibitem{yao2021imagetbad}
Z.~Yao {\em et~al.}, ``Imagetbad: A 3d computed tomography angiography image dataset for automatic segmentation of type-b aortic dissection,'' {\em Front. Physiol.}, vol.~12, 2021.

\bibitem{zhang2019real}
C.~Zhang, L.~Yang, L.~Xu, G.~Wang, and W.~Wang, ``Real-time editing of man-made mesh models under geometric constraints,'' {\em Computers \& Graphics}, vol.~82, pp.~174--182, 2019.

\bibitem{garcia2018sensitivity}
G.~Garc{\'\i}a-Isla {\em et~al.}, ``Sensitivity analysis of geometrical parameters to study haemodynamics and thrombus formation\,in\,the\,left atrial appendage,'' {\em Int J Num Method Biomed Eng}, vol.~34, no.~8, p.~e3100, 2018.

\bibitem{Schussnig2021b}
R.~Schussnig, D.~Pacheco, and T.~Fries, ``Robust stabilised finite element solvers for generalised newtonian fluid flows,'' {\em J Comput Phys}, vol.~442, p.~110436, 2021.

\bibitem{Schussnig2022c}
R.~Schussnig, D.~Pacheco, M.~Kaltenbacher, and T.-P. Fries, ``Semi-implicit fluid--structure interaction in biomedical applications,'' {\em Comput Methods Appl Mech Eng}, vol.~400, p.~115489, 2022.

\bibitem{Schussnig2021d}
R.~Schussnig, D.~Pacheco, and T.~Fries, ``Efficient split-step schemes for fluid-structure interaction involving incompressible generalised newtonian flows,'' {\em Comput Struct}, vol.~260, p.~106718, 2022.

\bibitem{DuenasPamplona2021atriumCFD}
J.~Dueñas-Pamplona, J.~G. Garc\'{i}a, J.~Sierra-Pallares, C.~Ferrera, R.~Agujetas, and J.~R. L\'{o}pez-M\'{i}nguez, ``A comprehensive comparison of various patient-specific {CFD} models of the left atrium for atrial fibrillation patients,'' {\em Computers Biol Med}, vol.~133, p.~104423, 2021.

\bibitem{Ranftl2021}
S.~Ranftl, T.~M{\"u}ller, U.~Windberger, W.~{von der Linden}, and G.~Brenn, ``{A {B}ayesian approach to blood rheological uncertainties in aortic hemodynamics},'' {\em Int J Numer Method Biomed Eng}, 2022.

\bibitem{ExaDG2020}
D.~Arndt, N.~Fehn, G.~Kanschat, K.~K.~M. Kronbichler, P.~Munch, W.~Wall, and J.~Witte, ``{ExaDG}: High-order discontinuous {G}alerkin for the exascale,'' in {\em Software for Exascale Computing - SPPEXA 2016-2019}, (Cham), pp.~189--224, Springer International Publishing, 2020.

\bibitem{dealII95}
D.~Arndt, W.~Bangerth, M.~Bergbauer, M.~Feder, M.~Fehling, J.~Heinz, T.~Heister, L.~Heltai, M.~Kronbichler, M.~Maier, P.~Munch, J.~Pelteret, B.~Turcksin, D.~Wells, and S.~Zampini, ``{The \texttt{deal.II} Library, Version 9.5},'' {\em J Numer Math}, 2023.

\bibitem{Fehn2018}
N.~Fehn, W.~A. Wall, and M.~Kronbichler, ``Robust and efficient discontinuous galerkin methods for under-resolved turbulent incompressible flows,'' {\em J Comput Physics}, vol.~372, pp.~667--693, 2018.

\bibitem{Fehn2020}
N.~Fehn, P.~Munch, W.~Wall, and M.~Kronbichler, ``{Hybrid multigrid methods for high-order discontinuous {G}alerkin discretizations},'' {\em J Comput Phys}, vol.~415, p.~109538, 2020.

\bibitem{viville2023meso}
P.~Viville, P.~Kraemer, and D.~Bechmann, ``Meso-skeleton guided hexahedral mesh design,'' {\em Comput Graph Forum}, p.~e14932, 2023.

\end{thebibliography}

\end{document}